\documentclass{sig-alternate-05-2015}
\usepackage{multirow}
\usepackage{epstopdf}
\usepackage{etoolbox}
\usepackage{amsmath}
\usepackage{tabularx}
\usepackage{multirow}
\usepackage{graphicx}
\usepackage{amssymb}
\usepackage{algorithm}
\usepackage{algorithmicx}
\usepackage{algpseudocode}
\usepackage{subcaption}
\usepackage{resizegather}

\usepackage{amsthm}

\newtheorem{lemma}{Lemma}
\newtheorem{proposition}{Proposition}

\newtheorem{remark}{Remark}

\patchcmd{\maketitle}{\@copyrightspace}{}{}{}
\begin{document}





%
\title{Classifier Risk Estimation under Limited Labeling Resources}

\numberofauthors{2} 
%
\author{
%
%
\alignauthor
Anurag Kumar\\
       \affaddr{Language Technologies Institute}\\
       \affaddr{Carnegie Mellon University}\\
       \affaddr{Pittsburgh, PA, USA}\\
       \email{alnu@andrew.cmu.edu}
\alignauthor
Bhiksha Raj\\
       \affaddr{Language Technologies Institute}\\
       \affaddr{Carnegie Mellon University}\\
       \affaddr{Pittsburgh, USA}\\
       \email{bhiksha@cs.cmu.edu}
}

\maketitle
\begin{abstract}
In this  paper we propose strategies for estimating performance of a classifier when labels cannot be obtained for the whole test set. The number of test instances which can be labeled is very small compared to the whole test data size.  The goal then is to obtain a precise estimate of classifier performance using as little labeling resource as possible. Specifically, we try to answer, how  to select a subset of the large test set for labeling such that the performance of a classifier estimated on this subset is as close as possible to the one on the whole test set. We propose strategies based on stratified sampling for selecting this subset. We show  that these strategies can reduce the variance in estimation of classifier accuracy by a significant amount compared to simple random sampling (over $\mathbf{65\%}$ in several cases). Hence, our proposed methods are much more precise compared to random sampling for accuracy estimation under restricted labeling resources.  The reduction in number of samples required (compared to random sampling) to estimate the classifier accuracy with only $1\%$ error is high as $\mathbf{60\%}$ in some cases. 
\end{abstract}
%

\vspace{-0.05in}
\vspace{-0.05in}
\section{Introduction}
The process of applying machine learning to solve a problem is usually a two phase process. The first phase, usually referred to as \emph{training} phase involves learning meaningful models which can properly explain a given \emph{training} data for the problem concerned. The next phase is the \emph{testing} phase where the goal is to evaluate the performance of these models on an unseen data set (\emph{test}) of the same problem. This step is necessary to understand the suitability of the applied machine learning algorithm in solving the concerned problem. It is also required to compare two different algorithms. Our interest in this work is on classification problems and hence the training phase involves training a classifier over the training data and the testing phase involves obtaining the accuracy of the classifier on any test set.  

The two-phase process described above usually requires labeled data in both phases. Labeling data is a tedious and expensive procedure, often requiring manual processing. In certain cases one might need specialized experts for labeling, an example would be labeling of medical data. This can further raise the cost of labeling. Although, the bulk of the machine learning solutions relies on supervised training of classifiers, there have been concrete efforts to reduce the dependence on labeled data for training phase by developing unsupervised and semi-supervised machine learning algorithms \cite{friedman2001}. However, irrespective of the method employed in training phase, the {\em testing} phase always requires labeled data to compute classifier accuracy. Given that labeling is costly, the general tendency is to use most of the available labeling resources for obtaining labeled training data to provide supervision for learning. This leaves us wondering about the best strategy to {\em evaluate} classifier performance under limited labeling resources. 

The answer to this problem is necessary as we move more and more towards big data machine learning; classifier evaluation on large datasets needs to be addressed along with classifier training. It is worth noting that this problem is completely different from cross validation or any such method employed to measure the goodness of classifier during \emph{training} phase. How the classifier is trained is immaterial to us, our goal is to accurately estimate the accuracy of a given trained classifier on a test set with as little labeling effort as possible. A trained classifier is almost always applied on a dataset which was never seen before and to estimate classifier performance on that dataset we require it to labeled. This is also the case when a classifier is deployed into some real world application where test data can be extremely large and labeling even a small fraction of it might be very difficult. Moreover, one might have to actively evaluate classifier as test data keeps coming in. All of these makes \emph{testing} phase important where labeled data is needed to evaluate classifier. Very little effort has been made to address the constraints posed by labeling costs during classifier evaluation phase. 

Some attempts have been made for unsupervised evaluation of multiple classifiers \cite{jaffe2014}, \cite{parisi2014}, \cite{platanios2014}, \cite{donmez2010}. All of these works try to exploit outputs of multiple classifiers and use them to either rank classifiers, estimate classifier accuracies or combine them to obtain a more accurate metaclassifier. Although, unsupervised evaluation sounds very appealing, these methods are feasible only if multiple classifiers are present. Moreover, assumptions such as conditional independence of classifiers in most cases and/or knowledge of marginal distribution of class labels in some cases need to be satisfied. In contrast, our focus is on the more general and practical case where the goal is to estimate the accuracy of a single classifier without the aid of any other classifier. The labeling resources are limited, meaning the maximum number of instances from the test data for which labels can be obtained is fixed and in general very small compared to the whole test set. The problem now boils down to sampling  instances for labeling  such that the accuracy estimated on the sampled set is a close approximation of true accuracy. The simple strategy, of course, is {\em simple random sampling} -- randomly drawing samples from the test set. This approach is, however, inefficient, and the variance of the accuracy estimated can be quite large. Hence, the fundamental question we are trying to answer is: can we do better than random sampling, where the test instances or samples to be labeled are selected from the whole test set?

The answer is Yes and the solution lies in \emph{Stratified Sampling} which is a well known concept in statistics \cite{cochran2007}. In stratified sampling the major idea is to divide the data into different strata and then sample a certain number of instances from each stratum. The statistical importance of this process lies in the fact that it usually  leads to reduction in the variance of estimated variable. To apply stratified sampling, two important question needs to answered: (1) How to stratify the data (\emph{Stratification Methods})?  (2) How to allocate the total sample size across different strata (\emph{Allocation Methods}) ? We answer these questions with respect to classifier accuracy estimation and evaluate the reduction in variance of estimated accuracy when stratified sampling is used instead of random sampling. Very few works have looked into sampling techniques for classifier evaluation \cite{bennett2010},\cite{druck2011},\cite{katariya2012},\cite{sawade2010}. \cite{bennett2010} and \cite{druck2011} also used stratification for estimating classifier accuracy. Both of these works showed that stratified sampling in general leads to better estimate of classifier accuracy for a fixed labeling budget. However,  several important aspects are missing in these works, such as theoretical study of the variance of the estimators, thorough investigation into stratification and allocation methods, effect of number of strata in stratification, and also evaluation of non-probabilistic classifiers. Other factors such as analysis of dependence of the variance on the true accuracy is also missing.    

There are several novel contributions of this work where we employ stratified sampling for classifier accuracy estimation under limited labeling resources. We establish variance relationships for accuracy estimators using both random sampling and stratified sampling. The variance relations not only allow us to analyze stratified sampling for accuracy estimation in theory but also allows to directly compare variances in different cases empirically, leading to a comprehensive understanding. We propose 2 strategies for practically implementing \emph{Optimal} allocation in stratified sampling. We show that our proposed novel iterative method for optimal allocation offers several advantages over the non-iterative implementation of optimal allocation policy. The most important advantage is more precise estimation with lesser labeling cost. On the stratification front, we employ panoply of stratification methods and analyze their effect on the variance of estimated accuracy. More specifically, we not only look into stratification methods well established in statistical literature of stratified sampling but also consider clustering methods for stratification which are not directly related to stratified sampling. Another related aspect studied here is the effect of the number of strata on the estimation of accuracy. We show the success of our proposed strategies on both probabilistic as well as non-probabilistic classifiers. The only difference for these two types of classifiers lies in the way we use classifier scores for stratification. We also empirically study the dependence of preciseness in accuracy estimation on the actual value of true accuracy. Put simply, we look into whether stratified sampling is more effective for a highly accurate classifier or for a classifier with not so high accuracy.

In this work, we use only  classifier outputs for stratification. This is not only simpler but also less restrictive compared to cases where the feature space of instances is used for stratification \cite{katariya2012}.  There are a number of cases where the feature space might be unknown due to privacy and intellectual property issues. For example online text categorization or multimedia event detection may not give us the exact feature representations used for the inputs. These systems usually just give confidence or probability outputs of the classifier for the input. Medical data might bring in privacy issues in gaining knowledge of the feature space. Our method based only on classifier outputs is much more general and can be easily applied to any given classifier. The rest of the paper is organized as follows; In Section 2, we formalize the problem and the follow it up different estimation methods in Section 3. In Section 4 we describe our experimental study and then put our final discussion and conclusions in Section 5.
\section{PROBLEM FORMULATION}
\vspace{-0.01in}
\label{sec:prfrm}
Let $\mathcal{D}$ be a dataset with $N$ instances where $i^{th}$ instance is represented by $\vec{x}_i$. We want to estimate the accuracy of a classifier $C$ on dataset $\mathcal{D}$. The score output of the classifier on $\vec{x}_i$ is $C(\vec{x}_i)$ and the label predicted by $C$ for $\vec{x}_i$ is $\hat{l}_i$. Let $a_i$ be instance specific correctness measure such that $a_i=1$ if $l_i=\hat{l}_i$, otherwise $a_i=0$. Then the true accuracy, $A$, of the classifier over $\mathcal{D}$ can be expressed by Eq \ref{eq:tracc}. 
\begin{equation}
\label{eq:tracc}
A = \frac{\sum_{i=1}^N a_i }{N}
\end{equation}
Eq \ref{eq:tracc} is nothing but the population mean of variable $a_i$ where $\mathcal{D}$ represents the whole population. To compute $A$, we need to know $l_i$ for all $i=1\,\,to\,\,N$. Our problem is to estimate the true accuracy $A$ of $C$ under constrained labeling resources, meaning only a small number of instances, $n$, can be labeled. Under these circumstances we expect to chose samples for labeling in an intelligent way such that the estimated accuracy is as precise as possible. Mathematically, we are interested in an unbiased estimator of $A$ with minimum possible variance for a given $n$. 
\section{ESTIMATION METHODS}
\subsection{Simple Random Sampling}
\label{sec:rndsmp}
The trivial solution for the problem described in Section \ref{sec:prfrm} is to randomly select $n$ instances or samples and ask for labels for these instances. This process is called simple random sampling which we will refer to as random sampling at several places for convenience. Then the correctness measure $a_i$ can be computed for these selected $n$ instances, using which we can obtain an estimate of $A$. The estimate of the accuracy is the mean of $a_i$ over the sampled set, $\hat{A}^r = \frac{\sum_{i=1}^n a_i }{n}$.
$\hat{A}^r$ is an unbiased estimator of $A$ and the variance of $\hat{A}^r$ is given by Eq \ref{eq:vrnd}.

\begin{align}
V(\hat{A}^r) = \frac{S^2}{n}, \,\,where\,\,
S^2 = \frac{\sum\limits_{i=1}^{N} (a_i - A)^2}{N-1} \label{eq:vrnd}
\end{align}
$S^2$ is the variance of $a_i$ over $\mathcal{D}$. The variance formula above will include a factor $1-\frac{n}{N}$ if sampling without replacement. For convenience we will assume sampling with replacement in our discussion and hence this term will not appear. The following lemma establishes the variance $S^2$ of $a_i$ in terms of $A$. 
\begin{lemma}
\label{lemma1}
$S^2$ for $a_i$ is given by $ S^2 = \frac{N}{N-1}  \cdot A(1-A)$
\end{lemma}
\let\qed\relax
\vspace{-0.1in}
\begin{proof}
Expanding the sum in definition of $S^2$ in Eq \ref{eq:vrnd}
\begin{align*}
S^2  = & \frac{1}{N-1} (\sum\limits_{i=1}^N a_i^2 + \sum\limits_{i=1}^N  A^2 - \sum\limits_{i=1}^N  2A a_i)\\
     = & \frac{1}{N-1} (N \cdot A - N \cdot A^2) =  \frac{N}{N-1} \cdot A(1-A)\\
\end{align*}
\end{proof}
\vspace{-0.25in}
The second line follows from the fact that $a_i \in \{0,1\}$, hence, $\sum_{i=1}^{N}a_i^2 = \sum_{i=1}^{N}a_i$ and $\sum_{i=1}^{N}a_i = N \cdot A$. 

Using Lemma \ref{lemma1} in Eq \ref{eq:vrnd} establishes the following result for variance of $\hat{A}^r$. 
\begin{proposition}
\label{thm:vrnd}
The variance of random sampling based estimator of accuracy $\hat{A}^r$, is given by $V(\hat{A}^r) = \frac{N\,A(1-A)}{(N-1)\,\,n}$.
\end{proposition}
Since $A$ is unknown, we need an unbiased estimate of $V(\hat{A}^r)$ for empirical evaluation of variance. An unbiased estimate of $S^2$ can be obtained from a sample of size $n$ by $s^2 = \frac{\sum_{i=1}^n (a_i - \hat{A}^r)^2}{n-1}$, \cite{cochran2007}. Clearly, $a_i$ here corresponds to correctness measure for instances in the sampled set. Following the steps in Lemma \ref{lemma1}, we can obtain
\begin{equation}
\label{eq:s2ub}
s^2 = \frac{n}{n-1} \cdot \hat{A}^r(1-\hat{A}^r)
\end{equation}
\begin{proposition}
\label{thm:vrndub}
The unbiased estimate of variance of accuracy estimator $\hat{A}^r$, is given by $v(\hat{A}^r) = \frac{\hat{A}^r(1-\hat{A}^r)}{n-1}$.
\end{proposition}
Theorem \ref{thm:vrndub} follows from Eq \ref{eq:s2ub}. The estimated accuracy becomes more precise as $n$ increases due to decrease in variance with $n$. The important question is, how can we achieve more precise estimation or in other words lower variance at a given $n$? To understand the answer to this question let us look at it a slightly different way. The question can be restated as how many instances should be labeled for a fairly good estimate of accuracy $A$ ?  

\begin{figure}[t]
\centering
\includegraphics[trim=1.0in 0.50in 0.7in 0.30in,width=0.48\columnwidth,height=1.0in]{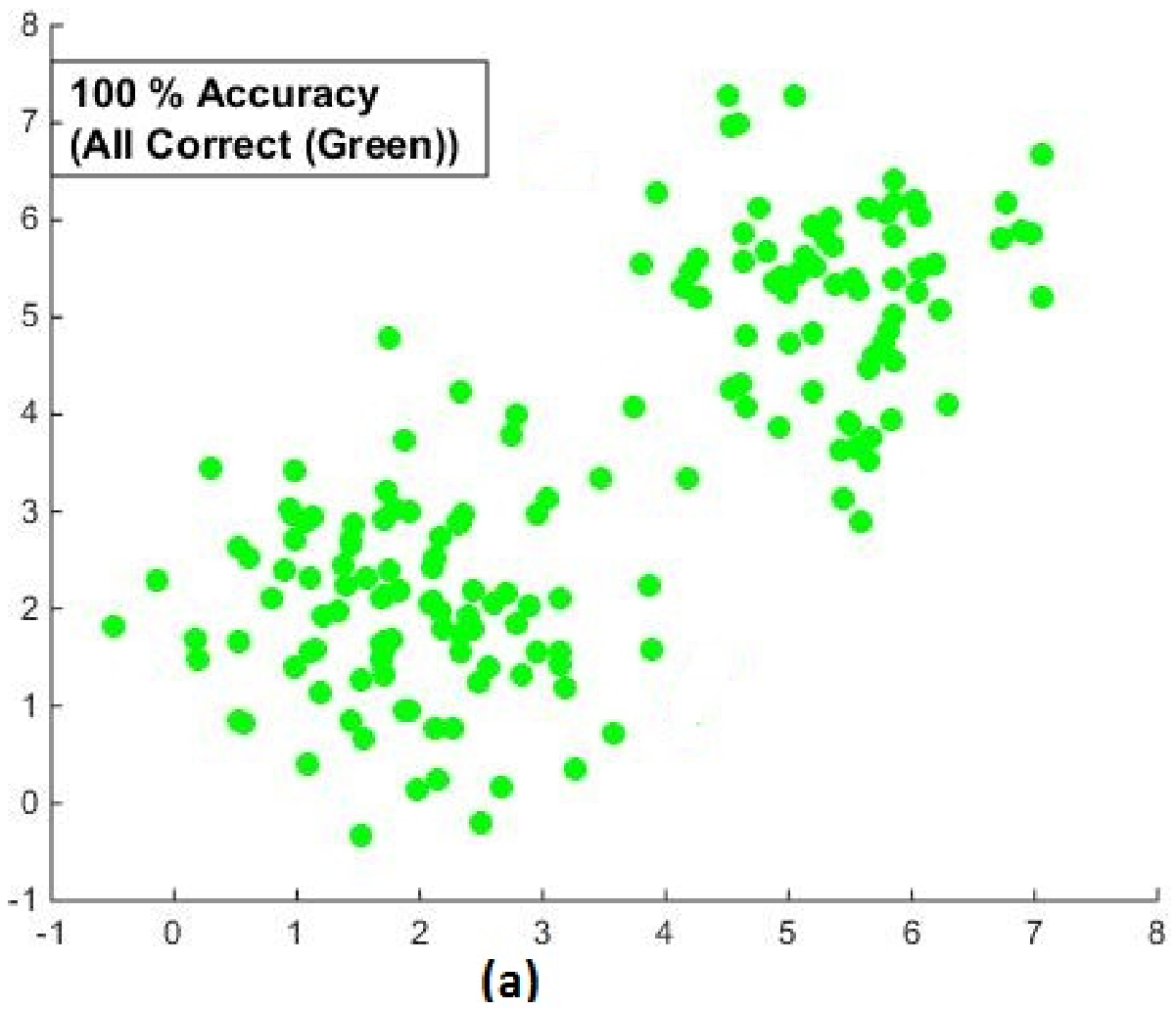}
\includegraphics[trim=1.0in 0.50in 0.7in 0.30in,width=0.48\columnwidth,height=1.0in]{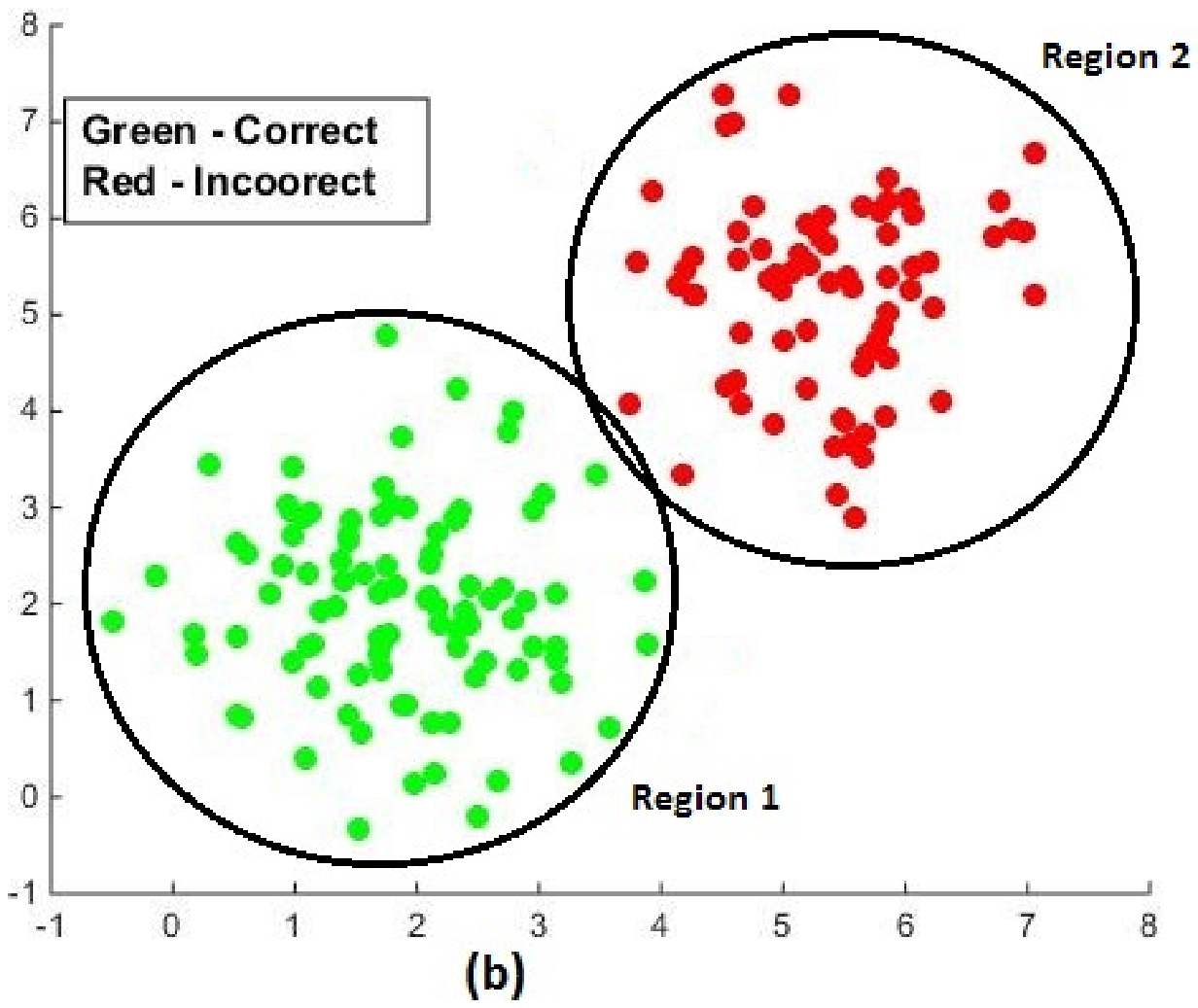}
\caption{Two Cases for Illustration}
\label{fig:exam}
\vspace{-0.15in}
\end{figure}

Consider Figure \ref{fig:exam}, where green points indicate instances for which $C$ correctly predicts labels ($a_i=1$). In Figure \ref{fig:exam}(a), the classifier is $100\%$ accurate. In this case a single instance is sufficient to obtain the true accuracy of classifier. Now consider Figure \ref{fig:exam}(b), where the classifier is $100\%$ accurate in \emph{Set 1} and $100\%$ incorrect in \emph{Set 2}. Thus, labeling $1$ instance from each set is sufficient to obtain true accuracy in that set and the overall accuracy is $A=\frac{1\times N_1+0\times N_2}{N}$. $N_1$ and $N_2$ are total number of points in sets 1 and 2 respectively. This leads us to the following general remark.
\begin{remark}
If $\mathcal{D}$ can be divided into $K$ ``pure'' sets, then true accuracy can be obtained by labeling $K$ instances only, where $1$ instance is taken from each set.  
\end{remark}
``Pure'' sets imply the classifier is either $100\%$ \emph{accurate} or $100\%$ \emph{inaccurate} in each set. In terms of the instance specific accuracy measure $a_i$, a ``pure'' set has either all $a_i=1$ or all $a_i=0$. This gives us the idea that if we can somehow divide the data into homogeneous sets then we can obtain a precise estimate of accuracy using very little labeling resources. The homogeneity is in terms of distribution of the values taken by $a_i$. The higher the homogeneity of a set the lesser the labeling resource we need for precise estimation of accuracy. Similarly, less homogeneous sets require more labeling resources. It turns out that this particular concept can be modeled in terms of a well known in statistics by the name of Stratified Sampling \cite{cochran2007}.
\subsection{Stratified Sampling}
Let us assume that the instances have been stratified into $K$ sets or strata. Let $\mathcal{D}_1,...,\mathcal{D}_K$ be those strata. The stratification is such that $\mathcal{D}_1 \cup \mathcal{D}_2 \cup ...  \cup \mathcal{D}_K = \mathcal{D}$ and $\mathcal{D}_j \cap \mathcal{D}_k = \emptyset, where,\,\,\,j \neq k, 1 \leq j \leq K, 1 \leq k \leq K$. All instances belong to only one stratum. The number of instances in strata $\mathcal{D}_k$ is $N_k$. Clearly, $\sum_{k=1}^K N_k = N$. The simplest form of stratified sampling is {\em stratified random sampling} in which samples are chosen randomly and uniformly from each stratum. If the labeling resource is fixed at $n$ then $n_k$ instances are randomly chosen from each stratum such that $\sum_{k=1}^K n_k=n$. In contrast to random sampling the estimate of accuracy by {\em stratified} random sampling is given by
\vspace{-0.15in}
\begin{equation}
\label{eq:stracc}
\hat{A}^s = \sum\limits_{k=1}^K \frac{N_k}{N} \hat{A}^r_k = \sum\limits_{k=1}^K W_k \hat{A}^r_k
\vspace{-0.2in}
\end{equation} 
where $\hat{A}^{r}_k = \frac{1}{n_k} \sum_{i=1}^{n_k} a_i $ and $W_k=N_k/N$ are the estimated accuracy in $k^{th}$ stratum and  weight of $k^{th}$ stratum respectively. The superscript $r$ denotes that random sampling is used to select instances within each stratum. On taking expectation on both sides of Eq \ref{eq:stracc}, it is straightforward to show that $\hat{A}^s$ is an \emph{unbiased estimator} of $A$. Under the assumption that instances are sampled independently from each stratum, the variance of $\hat{A}^s$ is $V(\hat{A}^s) = \sum\limits_{k=1}^K W_k^2 V(\hat{A}^r_k)$. Since sampling within a stratum is random, applying Theorem \ref{thm:vrnd} to each stratum leads to following result for the stratified sampling estimator. 
\begin{proposition}
\label{thm:vstr}
The variance of stratified random sampling estimator of accuracy, $\hat{A}^s$, is given by \hspace{1.0in} 
\begin{equation}
\label{eq:vstr}
V(\hat{A}^s) =  \sum\limits_{k=1}^K W_k^2 \frac{S^2_k}{n_k} = \sum\limits_{k=1}^K W_k^2 \frac{N_k\,A_k(1-A_k)}{(N_k-1)\,\,n_k}
\end{equation}
\end{proposition}
$S^2_k = \frac{N_k\,A_k(1-A_k)}{(N_k-1)}$ is the variance of the $a_i$'s in $k^{th}$ stratum. $A_k$ is the true accuracy in the $k^{th}$ stratum and clearly, $\sum_{k=1}^K W_kA_k=A$. 

Similarly, Theorem \ref{thm:vrndub} can be applied for each stratum to obtain an unbiased estimator of $V(\hat{A}^s)$.
\begin{proposition}
\label{thm:vstrub}
The unbiased estimate of variance of $\hat{A}^s$ is 
\begin{equation}
v(\hat{A}^s) = \sum\limits_{k=1}^K W_k^2 \frac{s^2_k}{n_k} =  \sum\limits_{k=1}^K W_k^2 \frac{\hat{A}^r_k(1-\hat{A}^r_k)}{(n_k-1)}
\end{equation}
\end{proposition}

The variance for stratified sampling is directly related to the two important questions posed for stratified sampling in the introduction of this paper. We answer the second question first which deals with methods for defining $n_k$ for each stratum. This allows a more systematic understanding of variance $V(\hat{A}^s)$ in different cases.
\subsection{Allocation Methods for Stratified Sampling}
\label{sec:alocMet}
We consider three different methods for distributing the available labeling resource $n$ among the strata. 
\subsubsection{Proportional (PRO) Allocation}
In proportional allocation the total labeling resource $n$ is allocated proportional to the weight of the stratum. This implies $n_k = W_k \times n$. Substituting this value in Eq \ref{eq:vstr}, the variance of $\hat{A}^s$ under proportional allocation, $V_{pro}(\hat{A}^s)$, is 
\begin{align}
\label{eq:vspro}
V_{pro}(\hat{A}^s) & = \frac{1}{n} \sum\limits_{k=1}^K W_k S^2_k \nonumber \\ 
& =  \frac{1}{n} \sum\limits_{k=1}^K W_k \frac{N_k\,A_k(1-A_k)}{(N_k-1)}
\end{align}
The unbiased estimate of $V_{pro}(\hat{A}^s)$ can be similarly obtained. Once the process of stratification has been done, stratified random sampling with  proportional allocation is fairly easy to implement. We compute $n_k$ and then sample and label $n_k$ instances from $k^{th}$ stratum to obtain an estimate of accuracy $A_k$.
\vspace{-0.05in}
\subsubsection{Equal (EQU) Allocation}
In Equal allocation the labeling resource is allocated equally among all strata. This implies $n_k = n/K$. Equal allocation is again straightforward to use for obtaining accuracy estimate. Under equal allocation the variance of estimator $\hat{A}^s$ is
\begin{align}
\label{eq:vsequ}
V_{equ}(\hat{A}^s) & = \frac{K}{n} \sum\limits_{k=1}^K W^2_k S^2_k \nonumber \\ 
& = \frac{K}{n} \sum\limits_{k=1}^K W^2_k \frac{N_k\,A_k(1-A_k)}{(N_k-1)}
\end{align}
\subsubsection{Optimal (OPT) Allocation}
Optimal allocation tries to obtain the most precise estimate of accuracy using stratified sampling, for a fixed labeling resource $n$. The goal is to minimize the variance in the estimation process. Optimal allocation factors in both the stratum size and variance within stratum for allocating resources. In this case the labeling resource allocated to each stratum is given by 
\begin{equation}
\label{eq:optalc}
n_k = n \frac{W_k S_k}{\sum_{k=1}^K W_k S_k}
\end{equation}
\vspace{-0.10in}
Using this value in Eq \ref{eq:vstr} the variance of $\hat{A}^s$ comes out as,
\begin{align}
\label{eq:vsopt}
V_{opt}(\hat{A}^s) & = \frac{\left(\sum\limits_{k=1}^K W_k S_k\right)^2}{n} \nonumber \\
 & = \frac{\left[ \sum\limits_{k=1}^K W_k \left(\frac{N_k\,A_k(1-A_k)}{(N_k-1)}\right)^{\frac{1}{2}} \right]^2}{n}
\end{align}

Thus, a larger stratum or a stratum with higher variance of $a_i$ or both is expected to receive more labeling resource compared to other strata. This variance based allocation is directly related to our discussion at the end of Sec \ref{sec:rndsmp}. We remarked that a stratum which is homogeneous in terms of accuracy and hence a low variance stratum requires very few samples for precise estimation of accuracy in that stratum and vice versa. Thus, the intuitive and mathematical explanation are completely in sync with each other.  

However, practical implementation of optimal allocation is not as straightforward as the previous two allocation methods. The true accuracies $A_k$'s and hence $S_k^2$ are unknown implying we cannot directly obtain values of $n_k$. We propose two methods for practical implementation of optimal allocation policy. 

\begin{algorithm}[t]
\caption{OPT-A1 Allocation}\label{alg:opt1}
\begin{algorithmic}[1]
\Procedure{OPT-A1}{$\mathcal{D}_1,...,\mathcal{D}_k$,$n_{ini}$}
\State \parbox[t]{\dimexpr\linewidth-\algorithmicindent} {Randomly Select and Label $n_{ini}$ instances from each stratum \strut}
\State \parbox[t]{\dimexpr\linewidth-\algorithmicindent} {Estimate $A_k$ and then $S^2_k$ for each strata (applying Eq \ref{eq:s2ub} for $k^{th}$ stratum)}
\State $n_{rem} = n - (K*n_{ini})$
\State \parbox[t]{\dimexpr\linewidth-\algorithmicindent} {Allocate $n_{rem}$ among strata using the estimated $S_k^2$ in Eq \ref{eq:optalc}}
\State \parbox[t]{\dimexpr\linewidth-\algorithmicindent} {Randomly sample again from each stratum according to above allocation}
\State Update estimates of $A_k$ and $S_k^2$ for all $k$
\EndProcedure
\end{algorithmic}
\end{algorithm}

In the first method, we try to obtain an initial estimate of all $A_k$ by spending some labeling resources in each stratum. This leads us to an algorithm that we refer to as \emph{OPT-A1}. The OPT-A1 method is shown in Algorithm \ref{alg:opt1}.  In the first step $n_{ini}$ instances are chosen randomly from each stratum for labeling. Then, an unbiased estimate of $S_k^2$ is obtained by using Eq \ref{eq:s2ub} for $k^{th}$ stratum. In the last step, these unbiased estimates are then used to allocate rest of the labeling resource ($n-K*n_{ini}$) according to optimal allocation policy given by Eq \ref{eq:optalc}. Then, we sample again from each stratum according to the amount of allocated labeling resources and then update estimates of $A_k$. 

\begin{algorithm}[t]
\caption{OPT-A2 Allocation}\label{alg:opt2}
\begin{algorithmic}[1]
\Procedure{OPT-A2}{$\mathcal{D}_1,...,\mathcal{D}_k$,$n_{ini}$,$n_{step}$}
\State \parbox[t]{\dimexpr\linewidth-\algorithmicindent} {Randomly Select and Label $n_{ini}$ instances from each stratum \strut}
\State Estimate $A_k$ and $S^2_k$ for each strata 
\State $n_{rem} = n - (K*n_{ini})$
\While{$n_{rem} > 0$}
\State $n_{curr} = min(n_{step},n_{rem})$
\State \parbox[t]{\dimexpr\linewidth-\algorithmicindent} {Allocate $n_{curr}$ among strata using current estimate of $S_k^2$ in Eq \ref{eq:optalc} \strut}
\State \parbox[t]{\dimexpr\linewidth-\algorithmicindent} {Select and label new instances from each stratum according to allocation of $n_{curr}$ in previous step \strut}
\State Update estimates of $A_k$ and $S_k^2$ for all $k$
\State $n_{rem} = n_{rem} - n_{curr}$
\EndWhile
\EndProcedure
\end{algorithmic}
\end{algorithm}

In theory, optimal allocation gives us the minimum possible variance in accuracy estimation. However, allocation of $n$ according to OPT-A1 depends heavily on initial estimates of $S_k^2$ in each stratum. If $n_{ini}$ is small we might not able to get a good estimate of $S_k^2$ which might result in an allocation far from true optimal allocation policy. On the other hand, if $n_{ini}$ is large we essentially end up spending a large proportion of the labeling resource in a uniform fashion which is same as equal allocation. This would reduce the gain in preciseness or reduction in variance we expect to achieve using optimal allocation policy. The optimal allocation in this case comes into picture for a very small portion ($n-K*n_{ini}$) of total labeling resource.

Practically, it leaves us wondering about value of parameter $n_{ini}$. To address this problem we propose another novel method for optimal allocation called OPT-A2. OPT-A2 is an iterative form of OPT-A1. The steps for OPT-A2 are described in Algorithm \ref{alg:opt2}. In OPT-A2 $n_{ini}$ is a small reasonable value. However, instead of allocating the remaining labeling resource in  the next step we adopt an adaptive formalism. In this adaptive formalism step we allocate a fixed $n_{step}$ labeling resource among the strata in each step. This is followed by an update in estimate of $A_k$ and $S_k^2$. The process is repeated till we exhaust our labeling budget. We later show that results for OPT-A2 are not only superior compared to OPT-A1 but also removes concerns regarding the right value of $n_{ini}$. We show that any small reasonable values of $n_{ini}$ and $n_{step}$ works well. 
\subsection{Comparison of Variances} 
In this Section we study the variance, $V(\hat{A}^s)$ of stratified accuracy estimate $\hat{A}^s$ in different cases. The first question that needs to answered is whether stratified variance $V(\hat{A}^s)$ is always lower than random sampling variance $V(\hat{A}^r)$ for a fixed $n$ or not. The answer depends on the sizes of strata $N_k$. We consider two cases; one in which all $1/N_k$ are small compared to $1$ and other in which it is not. 

\subsubsection{Case 1: $1/N_k$ negligible compared to 1}
This is the case we are expected to encounter in general for classifier evaluation and hence will be discussed in details. In this case, it can be easily established that, $V(\hat{A}^r) \geq V_{pro}(\hat{A}^s) \geq V_{pro}(\hat{A}^s)$ \cite{cochran2007}. For equal allocation no such theoretical guarantee can be established. We establish specific results below and compare variances of accuracy estimators for different cases. When needed, the assumption of $1/N_k << 1$ will be made.  

First, we consider the cases of $V(\hat{A}^{r})$ and $V_{pro}(\hat{A}^{s})$. If $1/N_k << 1$, then so is $1/N << 1$. Hence, $N_k/(N_k-1)$ and $N/(N-1)$ is almost 1. Under this assumption the difference between $V(\hat{A}^{r})$ and $V_{pro}(\hat{A}^{s})$ is
\begin{eqnarray}
\label{eq:pr}
\resizebox{0.85\columnwidth}{!}{$ V(\hat{A}^{r}) - V_{pro}(\hat{A}^{s}) = \frac{1}{n}[A(1-A) - \sum\limits_{k=1}^K W_k A_k(1-A_k)] $} \\
\resizebox{0.85\columnwidth}{!}{$  = \frac{1}{n}[\sum\limits_{k=1}^K W_k A_k^2 - A^2] = \frac{1}{n}\sum\limits_{k=1}^K W_k (A_k - A)^2$}
\end{eqnarray}

The second line uses the fact that $A=\sum W_kA_k$ and $\sum W_k = 1$. Eq \ref{eq:pr} implies that if the stratification is such that the accuracy of the strata are very different from each other, then the difference between $V(\hat{A}^{r})$ and $V_{pro}(\hat{A}^{s})$ is higher. This suggests that stratification which results in higher variance of $A_k$ will lead to higher reduction in the variance of accuracy estimator. A special case is when $A_k$ is same for all $k$. Then $A_k=A$ and in this case proportional allocation in stratified sampling will result in the same variance of estimated accuracy as simple random sampling. This implies that under this condition stratified sampling under proportional allocation is ineffective in improving the preciseness of accuracy estimation.  

For stratified sampling, $V_{opt}(\hat{A}^s)$ by definition is the minimum possible variance of $\hat{A}^s$ for a fixed $n$. At best we can expect $V_{pro}(\hat{A}^s)$ and $V_{equ}(\hat{A}^s)$ to attain $V_{opt}(\hat{A}^s)$. Consider the difference between $V_{pro}(\hat{A}^s)$ and $V_{opt}(\hat{A}^s)$. 
\begin{align}
& V_{pro}(\hat{A}^s) - V_{opt}(\hat{A}^s) =  \frac{1}{n} \left[\sum_{k=1}^K W_k S_k^2 - (\sum_{k=1}^K W_k S_k)^2 \right] \nonumber \\
& =  \frac{1}{n} [\sum_{k=1}^K W_k S_k^2 - S_M^2 ] = \frac{1}{n} \sum_{k=1}^K W_k (S_k - S_M)^2 \label{eq:stp2}
\end{align}
In the second step (Eq \ref{eq:stp2}), $S_M = \sum_{k=1}^K W_k S_k$ is the weighted mean of the $S_k$'s. The second equality in Eq \ref{eq:stp2} uses the definition of $S_M$ and the fact that $\sum_{k=1}^K W_k = 1$. 

From Eq \ref{eq:stp2} it is straightforward to infer that $V_{pro}(\hat{A}^s)$ and $ V_{opt}(\hat{A}^s)$ are equal \emph{if and only if} $S_k = S_M$. This basically implies that if stratification of $\mathcal{D}$ is such that $S_k$ is constant for all $k$ then the variance of the stratified accuracy estimator under proportional and optimal allocation are equal. Thus, proportional allocation is optimal in the sense of variance. 

Following the assumption of $1/N_k << 1$, $S_k=A_k(1-A_k)$. Let us assume that $S_k=S_c$ for all $k$, where $S_c$ is some constant value. $S_k=S_c=A_k(1-A_k)$ implies for a given $k$, the value of $A_k$ is one of the roots of the quadratic equation $y^2-y+S_c$. If all $A_k$ take the same root value, then from previous discussion we know $V(\hat{A}^{r}) = V_{pro}(\hat{A}^s)$. Constant $S_k$ also means $V_{pro}(\hat{A}^s) = V_{opt}(\hat{A}^s)$.  Hence, $V_{pro}(\hat{A}^s) = V_{opt}(\hat{A}^s) = V(\hat{A}^{r})$. This leads us to the following remark. 
\begin{remark}
\label{remk:2}
A stratification of $\mathcal{D}$ such that $A_k$ is same for all $k$ is the worst case stratification where the minimum possible variance of stratified estimator $\hat{A}^s$ is same as variance of random sampling accuracy estimator $\hat{A}^r$. 
\end{remark}
Thus, even though $S_k = Constant$ will lead to simpler proportional allocation achieving minimum possible variance, it is not a very favorable situation when compared to random sampling. We might end up in the situation of Remark \ref{remk:2}. Even if all $A_k$ do not take same value, $S_k=S_c$ for all $k$ implies the variance of $A_k$ will not be very high. Hence, under this condition the minimum possible variance $V_{opt}(\hat{A}^s) = V_{pro}(\hat{A}^s)$ for stratified sampling won't be significantly smaller than $V(\hat{A}^r)$. 

Now, assume $W_kS_k = S_{wc}$ for all $k$, where $S_{wc}$ is a fixed constant value. If $W_kS_k$ is constant then from Eq \ref{eq:vsequ}, $V_{equ}(\hat{A}^s) = \frac{K^2S_{wc}^2}{n}$. Also from Eq \ref{eq:vsopt}, $V_{opt}(\hat{A}^s) = \frac{K^2S_{wc}^2}{n}$. Hence, a stratification such that $W_kS_k$ is a constant implies equal allocation is as good as optimal allocation. Hence, if it can be ensured that $W_kS_k=Constant$, then the simpler equal allocation can substitute optimal allocation. 

Practical implementation of proportional and equal allocation methods are much simpler compared to optimal allocation where we need OPT-A1 or OPT-A2. In this Section apart from providing a comparison of variances in different cases, we looked into conditions under which proportional or equal allocation can be used as a substitute for optimal allocation giving same variance of estimator. For proportional allocation it did not turned out to be highly desirable because large reduction in variance compared to simple random sampling cannot be expected. 

Equal allocation seems to be a better option provided the condition of constant $W_kS_k$ is satisfied. However, this condition is important and we cannot blindly use equal allocation for any stratification of $\mathcal{D}$. This is due to the fact that unlike proportional and optimal it does not come with a theoretical guarantee that worst case variance will be same as simple random sampling. In fact in certain cases it can lead to a higher variance than simple random sampling. However, our empirical evaluation suggests that equal allocation works fairly well for a variety of stratification methods. Lastly, implementation of optimal allocation is not directly possible and it is possible that the empirical variance of optimal allocation becomes more than that of random sampling even if $1/N_k << 1$ is satisfied. However, using our proposed algorithms OPTA1 and OPTA2 it happens very rarely.
\subsubsection{Case 2: $1/N_k$ not negligible compared to 1}
In general, even for moderately sized dataset we are not expected to encounter this case. Hence, for simplicity we only briefly discuss this case and show that under this condition $V_{pro}(\hat{A}^s)$ and $V_{opt}(\hat{A}^s)$ need not always be less than $V(\hat{A}^r)$. Consider a specific case of stratification when all $A_k$ are equal. Hence, $A_k=A$ for all $k$. Now, the difference between $V(\hat{A}^r)$ and $V_{pro}(\hat{A}^s)$. 
\begin{equation}
\resizebox{0.88\columnwidth}{!}{$ V(\hat{A}^r) - V_{pro}(\hat{A}^s) =  \frac{NA(1-A)}{n(N-1)} - \sum_{k} W_k \frac{N_k A_k(1-A_k)}{n(N_k-1)} $}
\end{equation}
Using the fact that $A_k=A$
\begin{align*}
& V(\hat{A}^r) - V_{pro}(\hat{A}^s)= \frac{A(1-A)}{n} [ \frac{N}{N-1} - \sum_k \frac{N_k}{N} \frac{N_k}{N_k-1} ]\\
& = \frac{A(1-A)}{n} \left[ \sum_k \frac{N_k}{N-1} - \frac{N_k^2}{N(N_k-1)} \right]\\
& = \frac{A(1-A)}{n} \left[ \sum_k - \frac{N-N_k}{N(N-1)(N_k-1)} \right]\\
\vspace{-0.1in}
\end{align*}
Thus $V(\hat{A}^r) - V_{pro}(\hat{A}^s) < 0$. Hence, proportional stratified sampling gives higher variance than simple random sampling when all $A_k=A$. It is also possible to show that when $S_k^2$ is constant then it can lead to $V_{opt}(\hat{A}^s) = V_{pro}(\hat{A}^s) > V(\hat{A}^r)$. 
\subsection{Stratification Methods}
We now consider the other aspect of stratified sampling which is construction of strata. Let us denote the variable used for stratification by $z$ and let $f(z)$ be the density distribution of $z$. $z_i,\,\,i=1\,\,to\,\,N$ denotes the discrete values of stratification variable for instances in dataset $\mathcal{D}$. If the classifier outputs $C(\vec{x}_i)$ are probabilistic then we use $z_i = p(\hat{l}_i/\vec{x}_i)$, that is the stratification variable is the probability of the predicted class for $\vec{x}_i$. If the classifier scores are non-probabilistic and the predicted label is given by $\hat{l}_i = sign(C(\vec{x}_i))$, we use $z_i = |C(\vec{x}_i)|$, that is the magnitude of the classifier output. This particular stratification variable has been designed keeping in mind binary classifiers like support vector machines where scores of larger magnitude generally imply a greater level of confidence in the label assigned. These two approaches can be used as a general schema for extending the definition of the stratification variable for other types of classifier as well. 

The optimum stratification (in the sense of minimum variance) usually depends on the allocation policy \cite{sethi1963}\cite{dalenius1950}\cite{dalenius1951}. While relationships for optimum stratification for a given allocation method exist and can be solved by complicated iterative procedures, a large body of stratification literature consists of approximate methods for optimum stratification. 

We employ several known stratification methods for stratifying $\mathcal{D}$ using $z$. We also introduce use of clustering and simpler rule based methods which are usually not found in stratification literature. To estimate the density distribution $f(z)$ of the stratification variable using $z_i$'s, we use Kernel Density estimation methods \cite{friedman2001} with Guassian kernels. 

\textbf{cum} $\mathbf{\sqrt{f}}$ (SQRT): This method proposed in \cite{dalenius1959} is perhaps the most popular and widely used method for stratification. The method has been designed for optimum allocation policy. The simple rule is to divide the cumulative of $\sqrt{f(z)}$ into equal intervals. The points of stratification, $z^s_1 < z^s_2 < .. < z^s_{K-1}$, correspond to the boundary points corresponding to these intervals. The $k^{th}$ stratum consists of the set of instances for which $z$ lies between $z^s_{k-1}$ and $z^s_{k}$. $z^s_0$ and $z^s_K$ can be set as $max$ and $min$ of $z$. 

\textbf{cum} $\mathbf{f^{\frac{1}{3}}}$ (CBRT): This method is same as the SQRT except that the cube root of $f(z)$ is used in place of square root \cite{singh1971}. The derivation of SQRT method makes an assumption that stratification and estimation variables are same which is usually not the case. CBRT was proposed keeping in mind that stratification variable ($z$) is in practice different from estimation variable ($a$) and a regression model was assumed in deriving this method. \cite{thomsen1976} argues in favor of CBRT if proportional allocation is to be used. 

\textbf{Weighted Mean}(WTMN): In this method the key idea is to to make the weighted mean of the stratification variable constant \cite{hansen1953}. It is much simpler compared to the previous 2 methods and was proposed earlier to the previous two methods. 

All of the previous methods try to approximate optimum stratification. These methods (SQRT and CBRT) work well if the stratification variable and estimation variable are highly correlated \cite{serfling1968}\cite{andersonimplications}. In more generic settings such as ours, no such assumption can be made for the stratification and estimated variable. Hence, we propose to introduce other techniques as well, which while not tailor-made for stratified sampling, can nevertheless serve as a way for stratification.

\textbf{Clustering Methods}: Clustering is one of the simplest ways to group the data $\mathcal{D}$ into different strata. We use K-means(KM)  and Gaussian Mixture Models (GMM) based clustering to construct strata using $z$.

\textbf{Simple Score Based Partitioning}: The stratification variable $z$ is obtained from classifier scores and we propose two simple partitioning methods. The first one is called EQSZ  (Equal Size) in which the instances in $\mathcal{D}$ are first sorted according to the stratification variable. Starting from the top, each stratum takes away an equal number $N/K$ of instances. It is expected that variation of $z$ within each strata will be small. We call the other method as EQWD (Equal Width). In this case the range of $z$ for $\mathcal{D}$ ($r = max(z) - min(z)$) is divided into sub-ranges of equal width. The points of stratification are $z^s_k = min(z)+rk/K\,\,, k=1\,\,to\,\,K$. $z^s_0=min(z)$ is used in this case.
\section{EXPERIMENTS AND RESULTS}
The variance of stratified sampling depends on three important factors, Allocation Method, Stratification Method and number of strata. We perform a comprehensive analysis of all of these factors. Each allocation method is applied on all $7$ stratification methods. We vary the number of strata from $2$ to $10$ to study the effect of $K$. Overall, this results in large number of experiments and we try to present the most informative results for each case in the paper.  

We use three different dataset in our study. The first one, which is smallest of the three is the \emph{News20 binary} dataset. It is the 2 class form of the text classification UCI News20 dataset \cite{keerthi2005}. It consist of a total of around $20000$ instances. We use $4000$ randomly selected instances for training a logistic regression classifier and the rest are used as test set $\mathcal{D}$ for which the classifier accuracy needs to be estimated. The second one is the \emph{epsilon} dataset from the Pascal Large Scale Challenge \cite{pascalC}. It contains $0.5$ \emph{million} instances of which we use a randomly selected $50,000$ for training a linear SVM. The remaining $0.45$ million instances are used as the test set $\mathcal{D}$. The third is the two-class form of the \emph{rcv1} text categorization dataset which is the largest of the three datasets \cite{lewis2004}. The test set $\mathcal{D}$ consists of around $0.7$ \emph{million} instances. A logistic regression classifier is trained on the training set. We use the LIBLINEAR \cite{fanliblinear} package for training all classifiers. All data are available for download from the LIBLINEAR website. Experiments on the three datasets together contain sufficient variation to study different aspects of accuracy estimation. 

We will quantify our results in two ways. The first is the ratio of the variance of the stratified accuracy estimator to a random sampling estimator at a given $n$, VR=$V(\hat{A}^s)/V(\hat{A}^r)$. Clearly, unbiased estimates of $V(\hat{A}^r)$ and $V(\hat{A}^s)$ are used to measure VR. Ideally VR should be less than 1; the lower it is the better it is. The second measure deals with absolute error (AE) percentage in estimating accuracy. Specifically, we look at the AE vs $n$ plot and observe the amount of labeling resource required to achieve just $1\%$ absolute error in accuracy estimates. We focus on $\%$ reduction if any in required $n$ to achieve $1\%$ error when using $\hat{A}^s$ in place of $\hat{A}^r$. All experiments are repeated for $3000$ runs and the variance and error terms are means over these runs. Hence, we will use MVR and MAE to refer to mean variance ratio and mean absolute error respectively. 

\subsection{Proportional Allocation}
\begin{figure}[t]
  \begin{subfigure}[b]{0.49\columnwidth}
    \includegraphics[trim=1.0in 0.30in 0.7in 1.0in,width=1.0\columnwidth,height=1.0in]{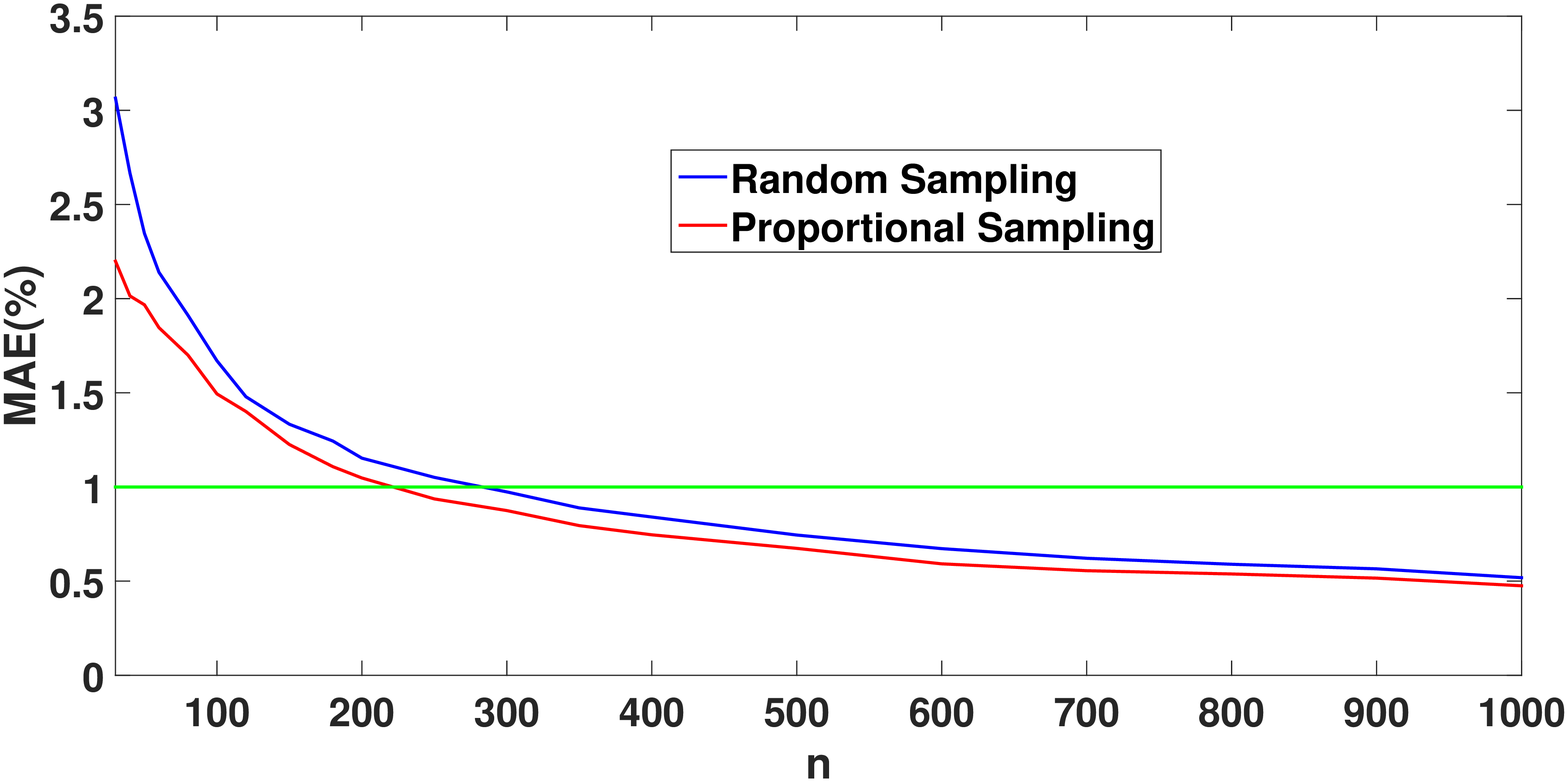}
    \caption{rcv1, $K=10$, EQWD}
    \label{fig:p1a}
  \end{subfigure}
  \begin{subfigure}[b]{0.49\columnwidth}
   \includegraphics[trim=1.0in 0.30in 1.0in 1.0in,width=1.0\columnwidth,height=1.0in]{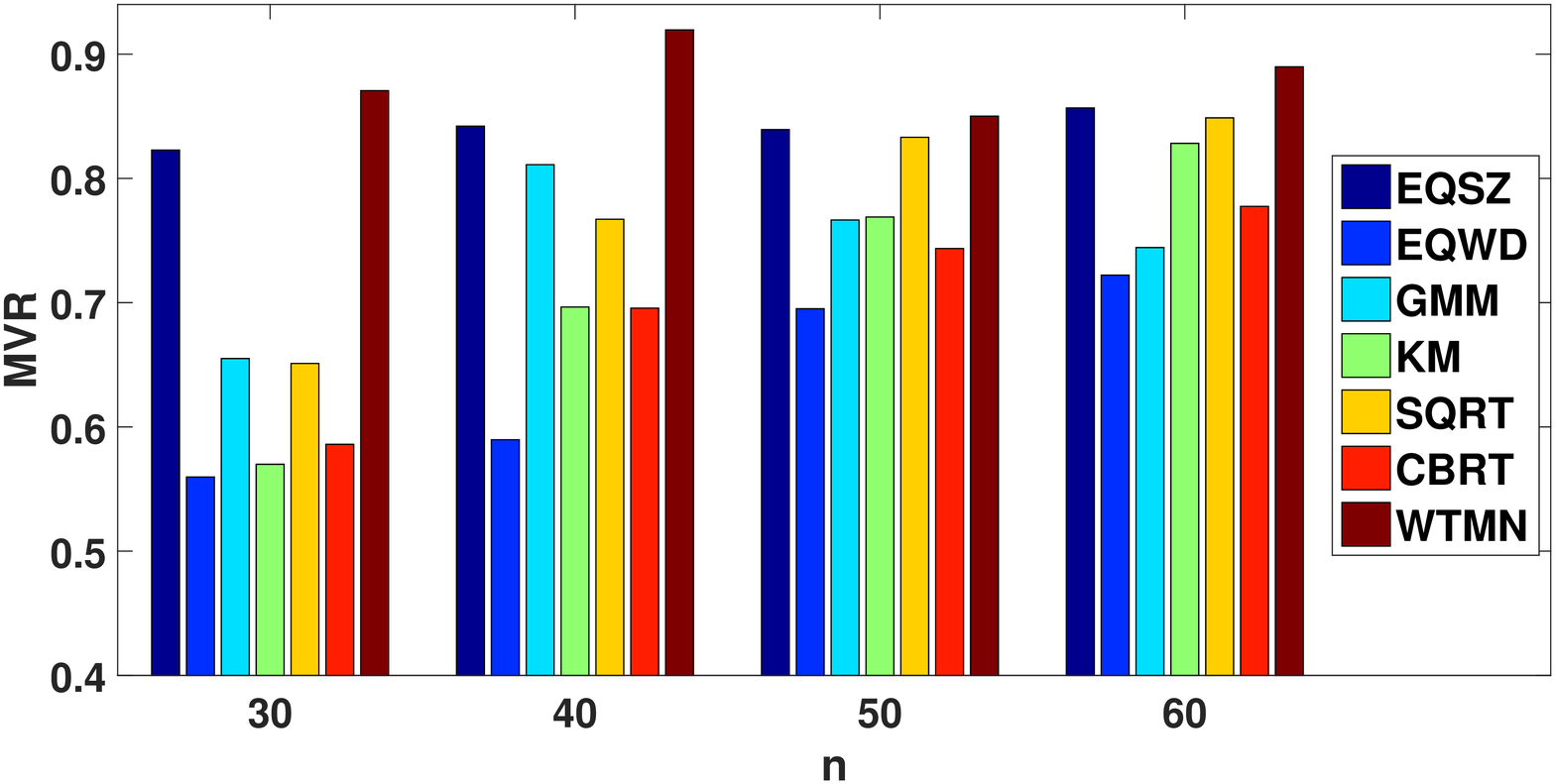}
    \caption{rcv1, $K=10$}
    \label{fig:p1b}
  \end{subfigure}
  \begin{subfigure}[b]{0.49\columnwidth}
    \includegraphics[trim=1.0in 0.40in 0.7in 0.30in,width=1.0\columnwidth,height=1.0in]{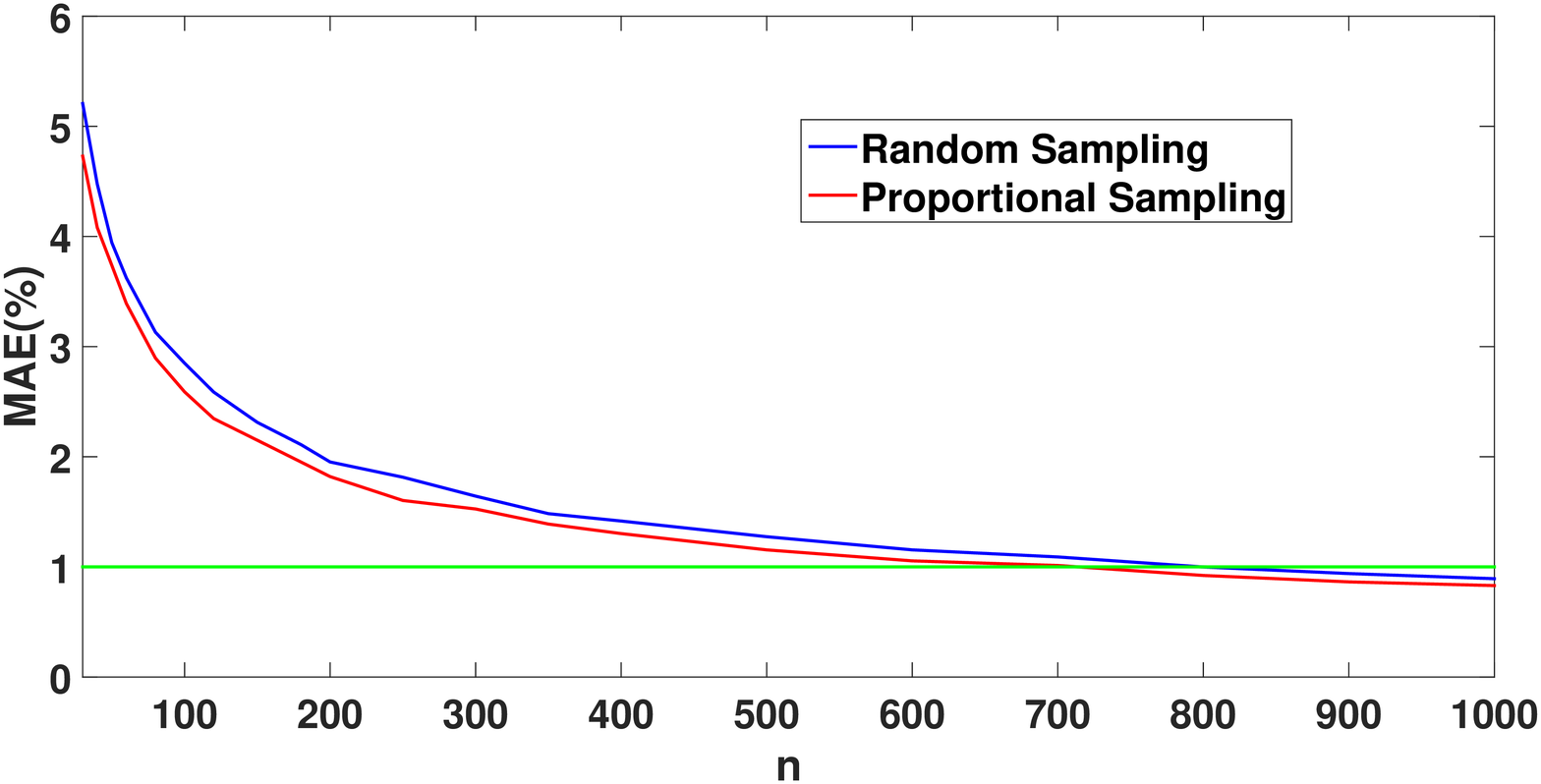}
    \caption{epsilon, $K=10$, EQSZ}
    \label{fig:p2a}
  \end{subfigure}
  \begin{subfigure}[b]{0.49\columnwidth}
   \includegraphics[trim=1.0in 0.40in 1.0in 0.30in,width=1.0\columnwidth,height=1.0in]{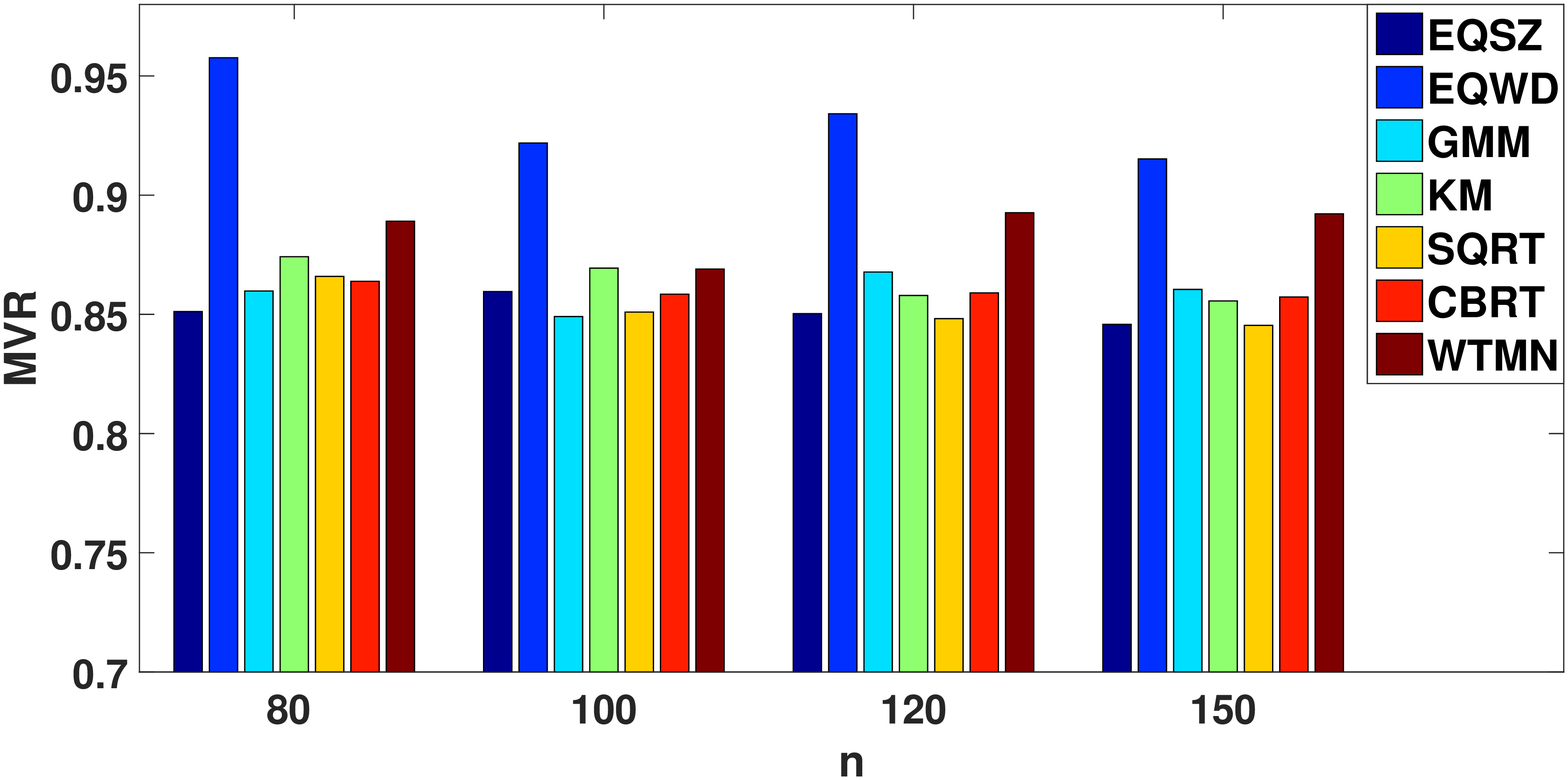}
    \caption{epsilon, $K=10$}
    \label{fig:p2b}
  \end{subfigure}
    \begin{subfigure}[b]{0.49\columnwidth}
    \includegraphics[trim=1.0in 0.50in 0.7in 0.25in,width=1.0\columnwidth,height=1.0in]{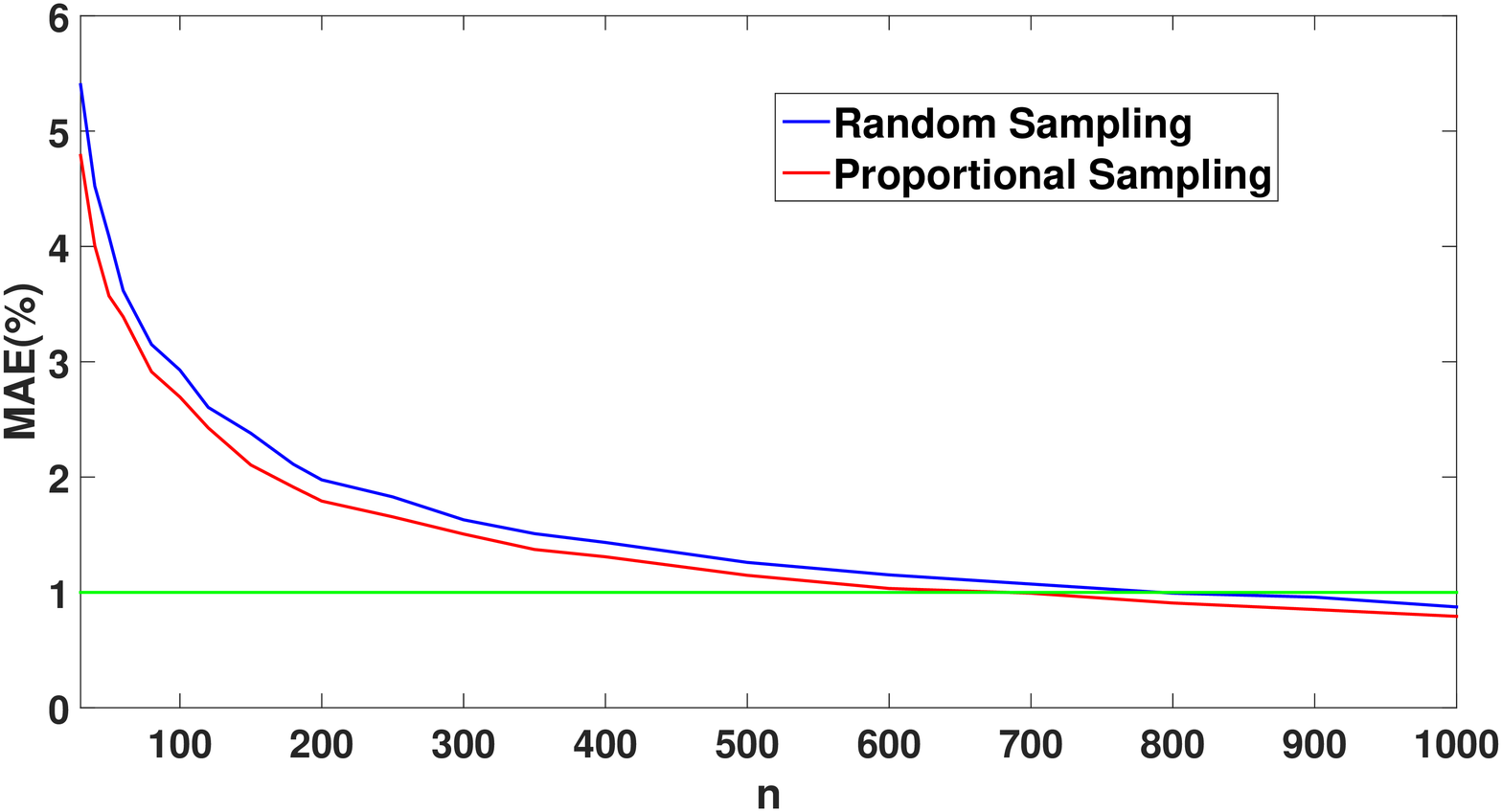}
    \caption{news20, K=10, KM}
    \label{fig:p3a}
  \end{subfigure}
  \begin{subfigure}[b]{0.49\columnwidth}
   \includegraphics[trim=0.8in 0.50in 1.0in 0.25in,width=1.0\columnwidth,height=1.0in]{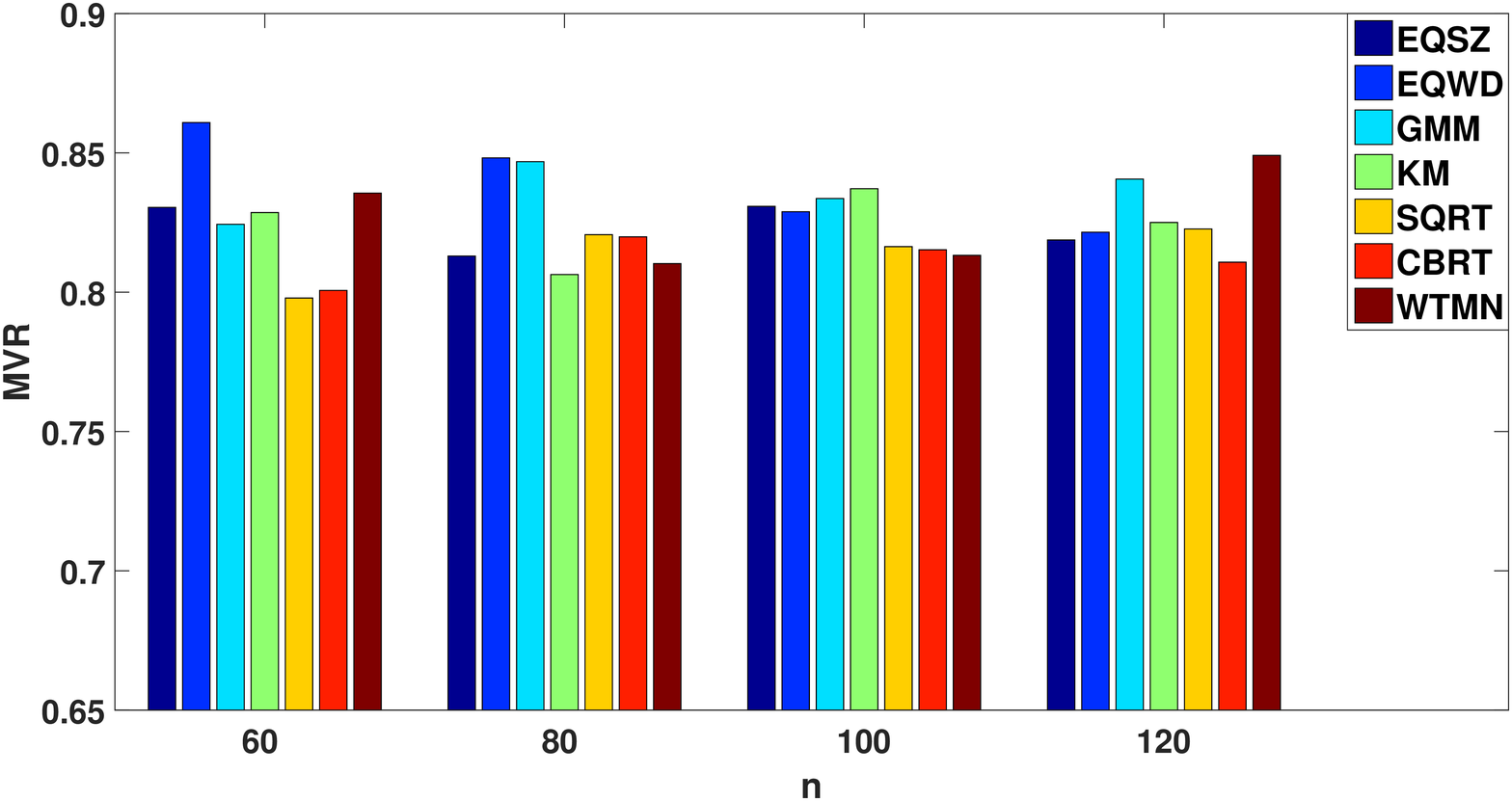}
    \caption{news20, $K=10$}
    \label{fig:p3b}
  \end{subfigure}
  \caption{Proportional Stratified Sampling}
\end{figure}

Figure \ref{fig:p1a} shows the MAE vs. n using EQWD stratification and $K=10$ for the \emph{rcv1} dataset. The number of labeled instances required to achieve a $1\%$ error in accuracy estimation goes down from $284$ in random sampling to $218$. This is about $\mathbf{23\%}$ reduction in labeling resources. Figure \ref{fig:p1b} shows the MVR values for each stratification method at different $n$. We can observe that EQWD is in general better compared to other methods leading to about $\mathbf{40-45\%}$ reduction in variance for some cases. WTMN is the worst showing only about $10\%$ reduction in variances. The lower $n$ values for which results are presented in Figure \ref{fig:p1b} and in subsequent figures, are in general more interesting cases. The difference between different methods are more visible and needs to be looked into carefully at lower $n$. 
\begin{figure}[t]
  \begin{subfigure}[b]{0.49\columnwidth}
    \includegraphics[trim=1.0in 0.30in 0.7in 1.0in,width=1.0\columnwidth,height=1.0in]{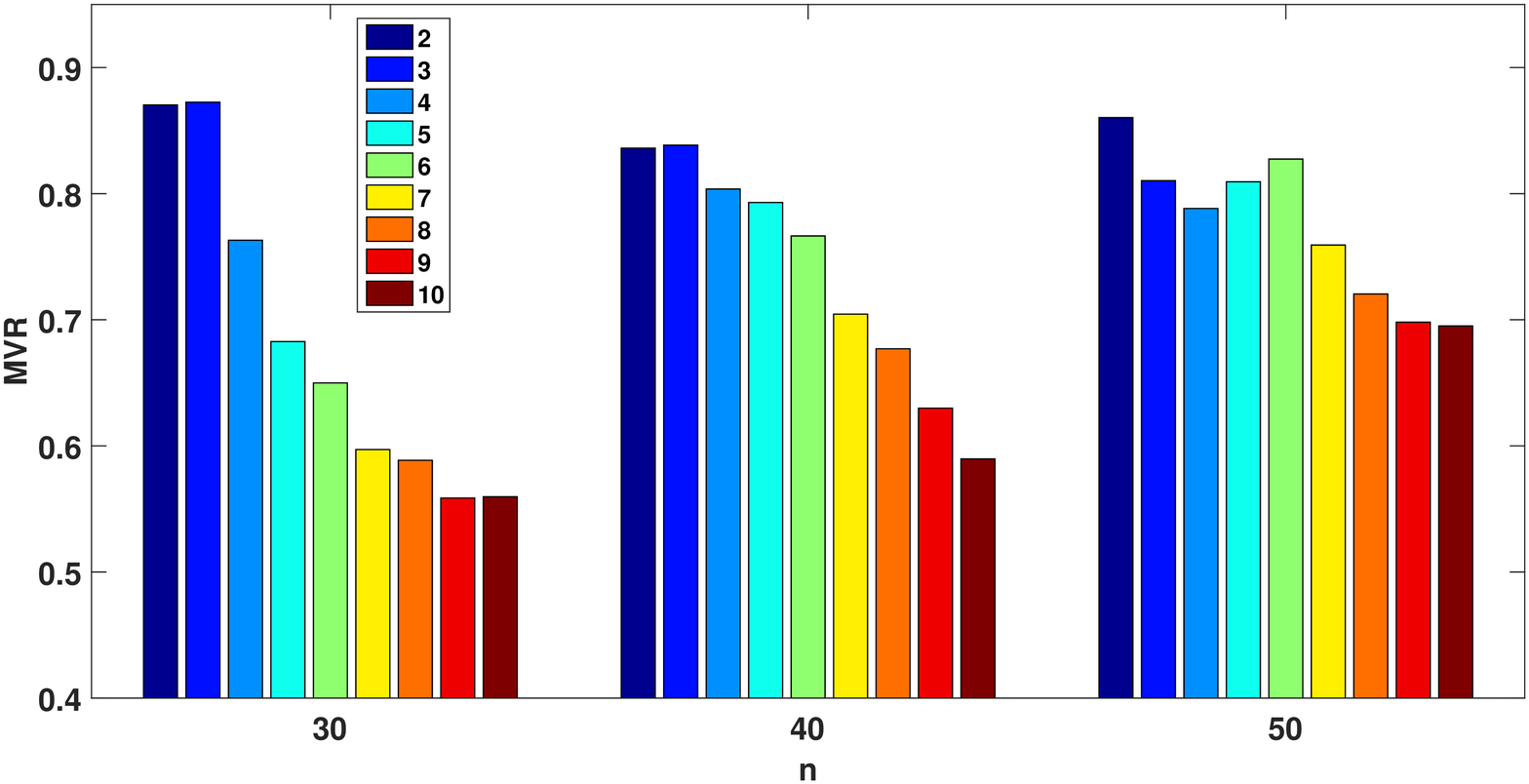}
    \caption{rcv1, EQWD, Proportional}
    \label{fig:pk1a}
  \end{subfigure}
  \begin{subfigure}[b]{0.49\columnwidth}
   \includegraphics[trim=1.0in 0.30in 1.0in 1.0in,width=1.0\columnwidth,height=1.0in]{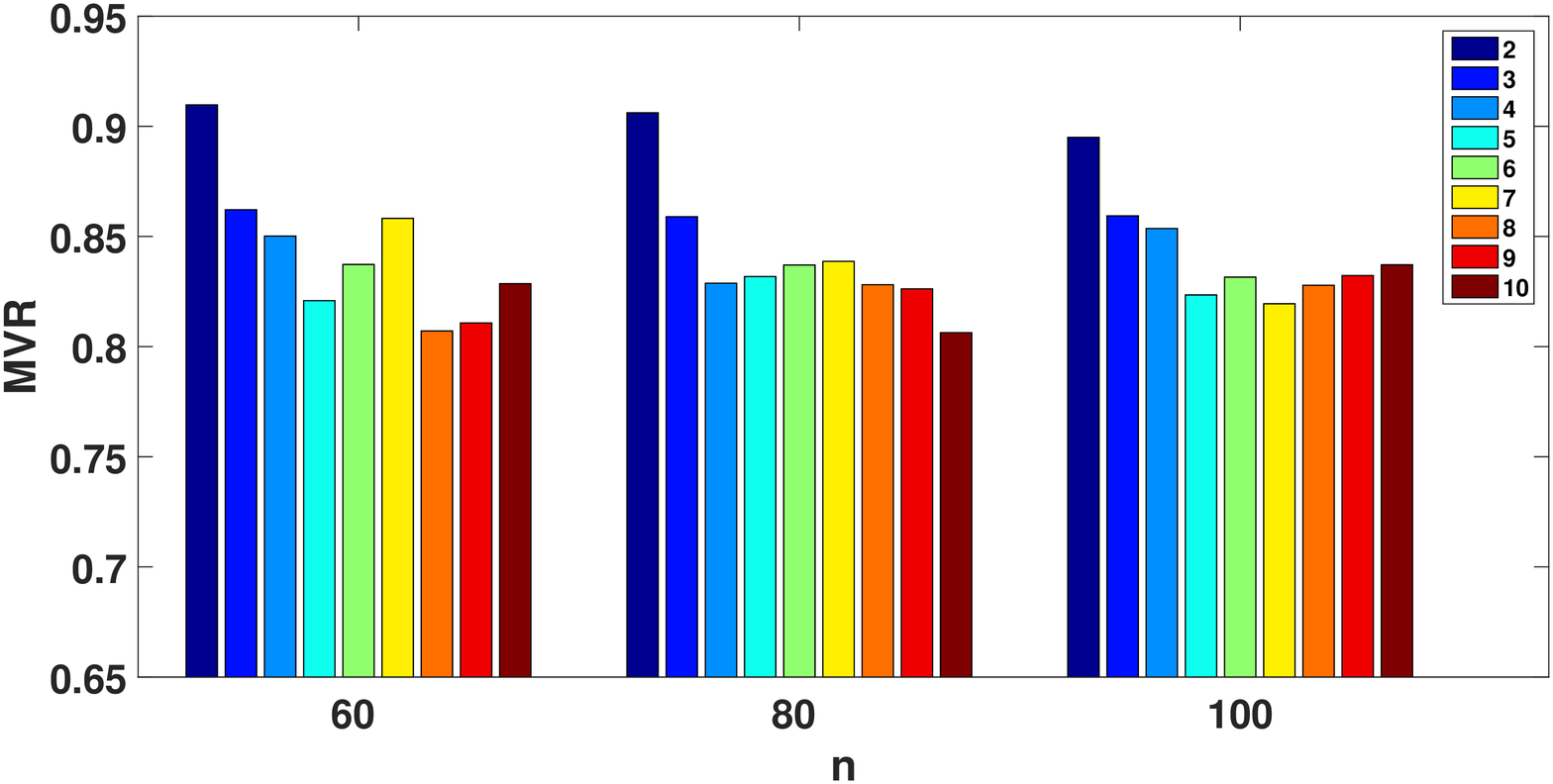}
    \caption{news20, KM, Proportional}
    \label{fig:pk1b}
  \end{subfigure}
  \caption{MVR Variation With K for Proportional}
	\vspace{-0.15in}  
\end{figure}

The results for the epsilon dataset are shown in Figure \ref{fig:p2a} and \ref{fig:p2b}. EQSZ stratification is used in Fig \ref{fig:p2a}. The reduction in labeling resources for $1\%$ error in accuracy estimation is about $12.5\%$. This is can be attributed to the fact that for this dataset the maximum reduction in variance with proportional stratified sampling is in general less than $20\%$. EQSZ performs only marginally better than other methods such as SQRT and CBRT. Figures \ref{fig:p3a} and \ref{fig:p3b} show results for the \emph{news20} dataset. About $16\%$ reduction in labeling resource can be observed for KM stratification method shown in Figure \ref{fig:p3a}. Although on average across different $n$ and $K$, $KM$ is slightly better than other methods it does not always dominate and SQRT and CBRT work almost as well.

The variation of MVR with $K$ for rcv1 and news20 is shown Figure \ref{fig:pk1a} and \ref{fig:pk1b} respectively. Increasing $K$ does not necessarily leads to better results. However, the general trend is that once $K$ is large enough major variation in MVR values cannot be expected. Hence, the parameter $K$ is important but setting it to fixed reasonable value which will lead to good estimation of accuracy does not appear to be a hard problem. The trend is same for epsilon dataset and hence not shown here for brevity. 

\subsection{Equal Allocation}
\begin{figure}[t]
  \begin{subfigure}[b]{0.49\columnwidth}
    \includegraphics[trim=1.0in 0.30in 0.7in 1.0in,width=1.0\columnwidth,height=1.0in]{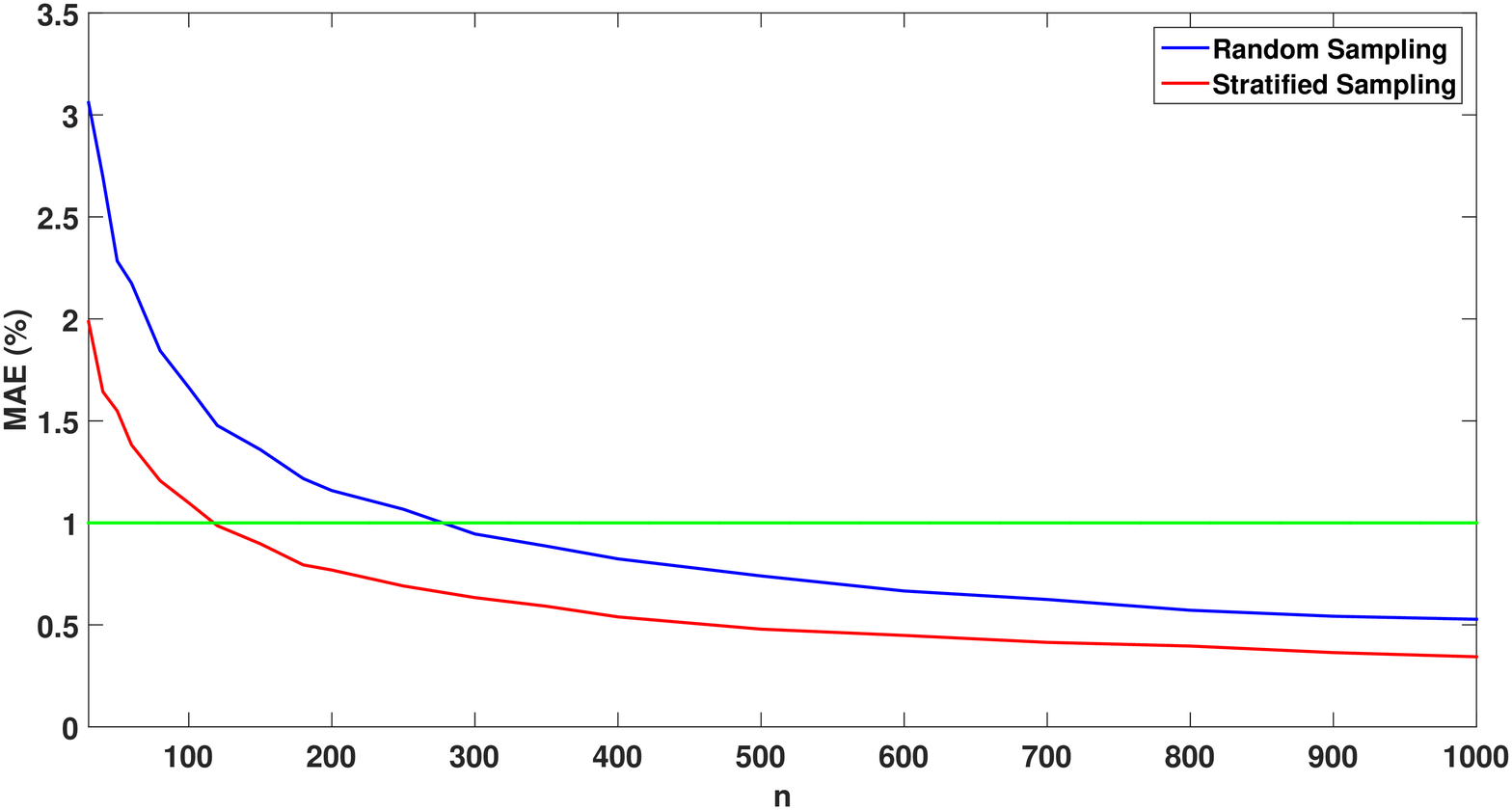}
    \caption{rcv1, $K=10$, KM}
    \label{fig:e1a}
  \end{subfigure}
  \begin{subfigure}[b]{0.49\columnwidth}
   \includegraphics[trim=1.0in 0.30in 1.0in 1.0in,width=1.0\columnwidth,height=1.0in]{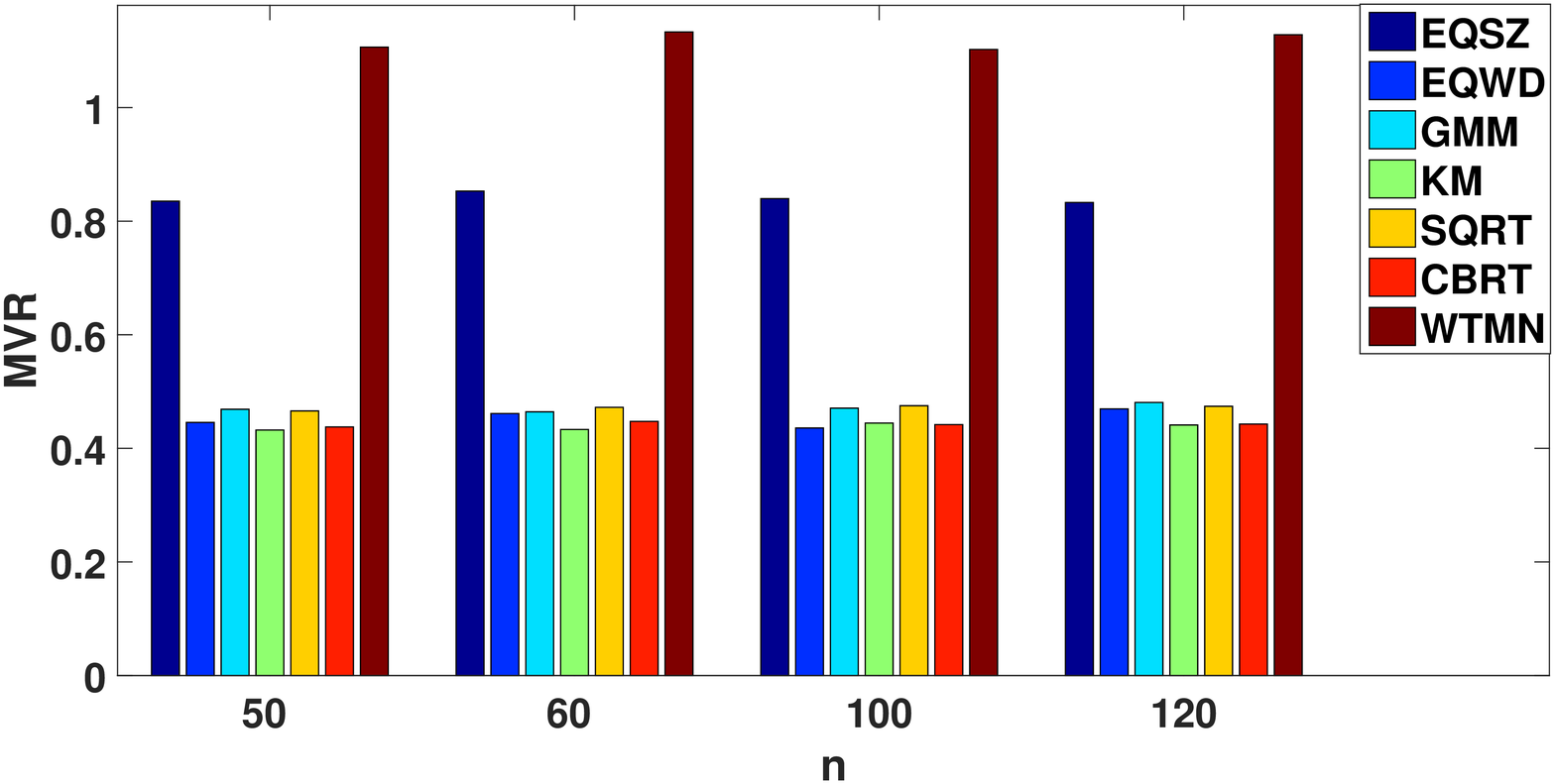}
    \caption{rcv1, $K=10$}
    \label{fig:e1b}
  \end{subfigure}
  \begin{subfigure}[b]{0.49\columnwidth}
    \includegraphics[trim=1.0in 0.40in 0.7in 0.30in,width=1.0\columnwidth,height=1.0in]{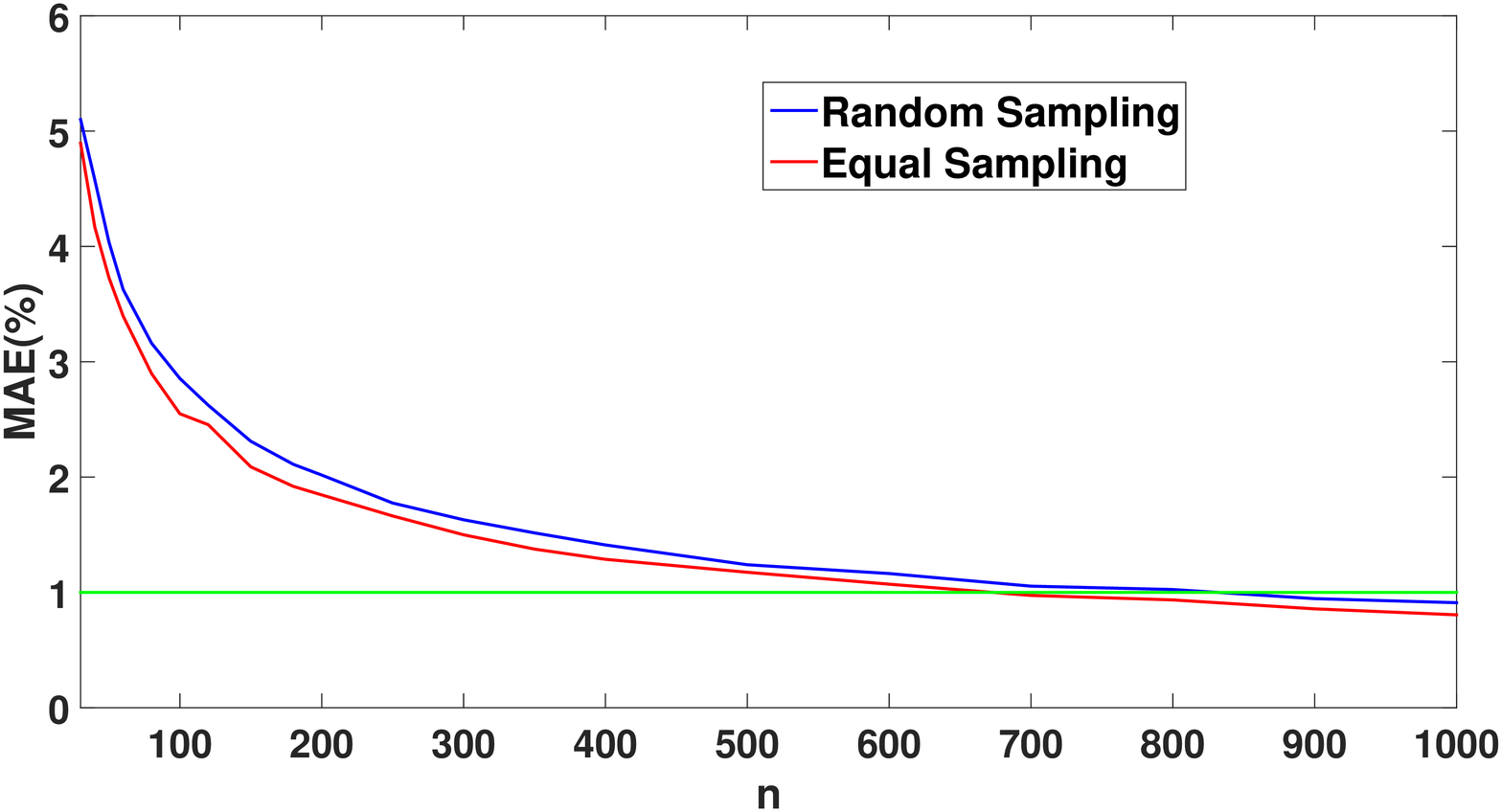}
    \caption{epsilon, $K=10$, EQSZ}
    \label{fig:e2a}
  \end{subfigure}
  \begin{subfigure}[b]{0.49\columnwidth}
   \includegraphics[trim=1.0in 0.40in 1.0in 0.30in,width=1.0\columnwidth,height=1.0in]{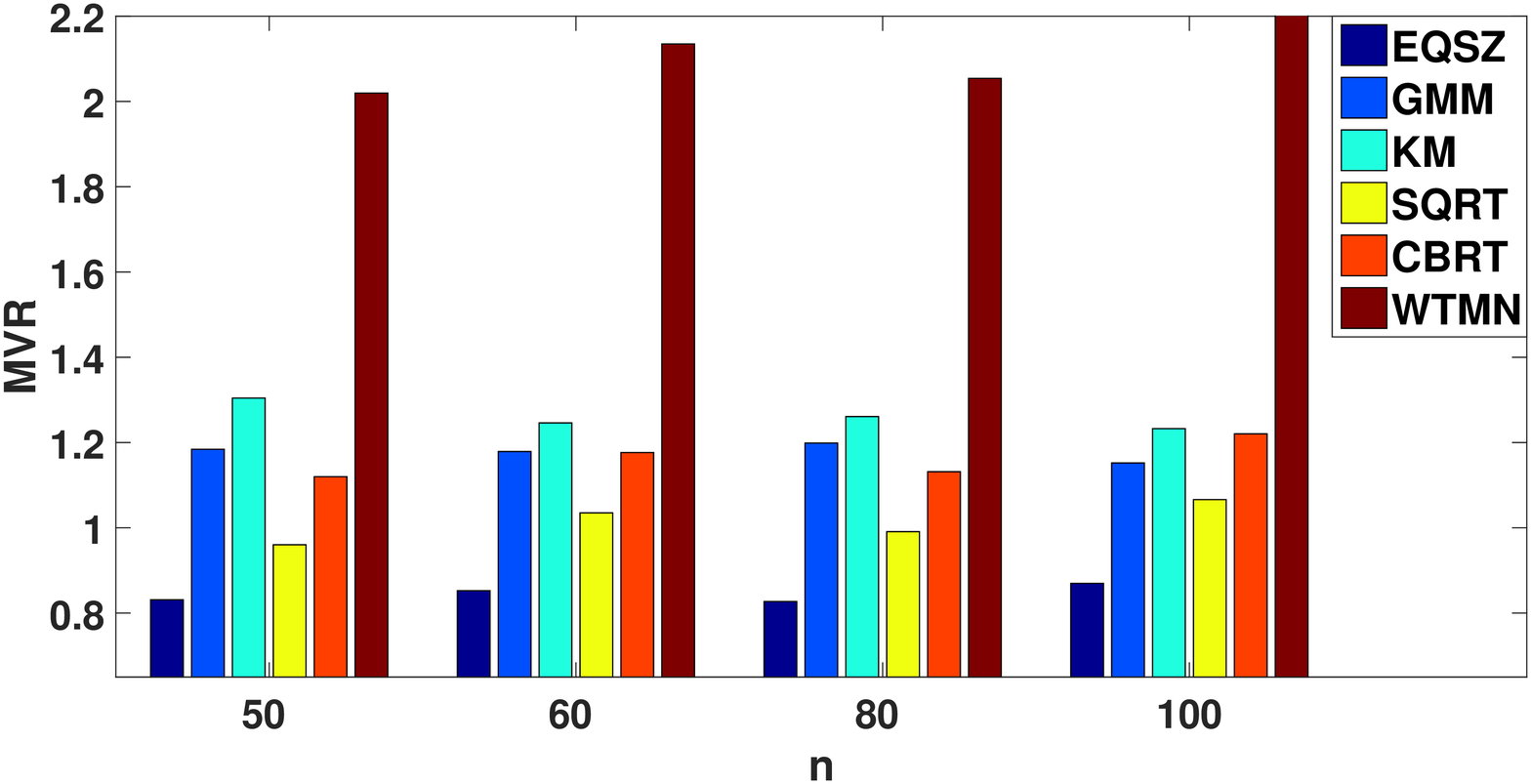}
    \caption{epsilon, $K=10$}
    \label{fig:e2b}
  \end{subfigure}
    \begin{subfigure}[b]{0.49\columnwidth}
    \includegraphics[trim=1.0in 0.50in 0.7in 0.25in,width=1.0\columnwidth,height=1.0in]{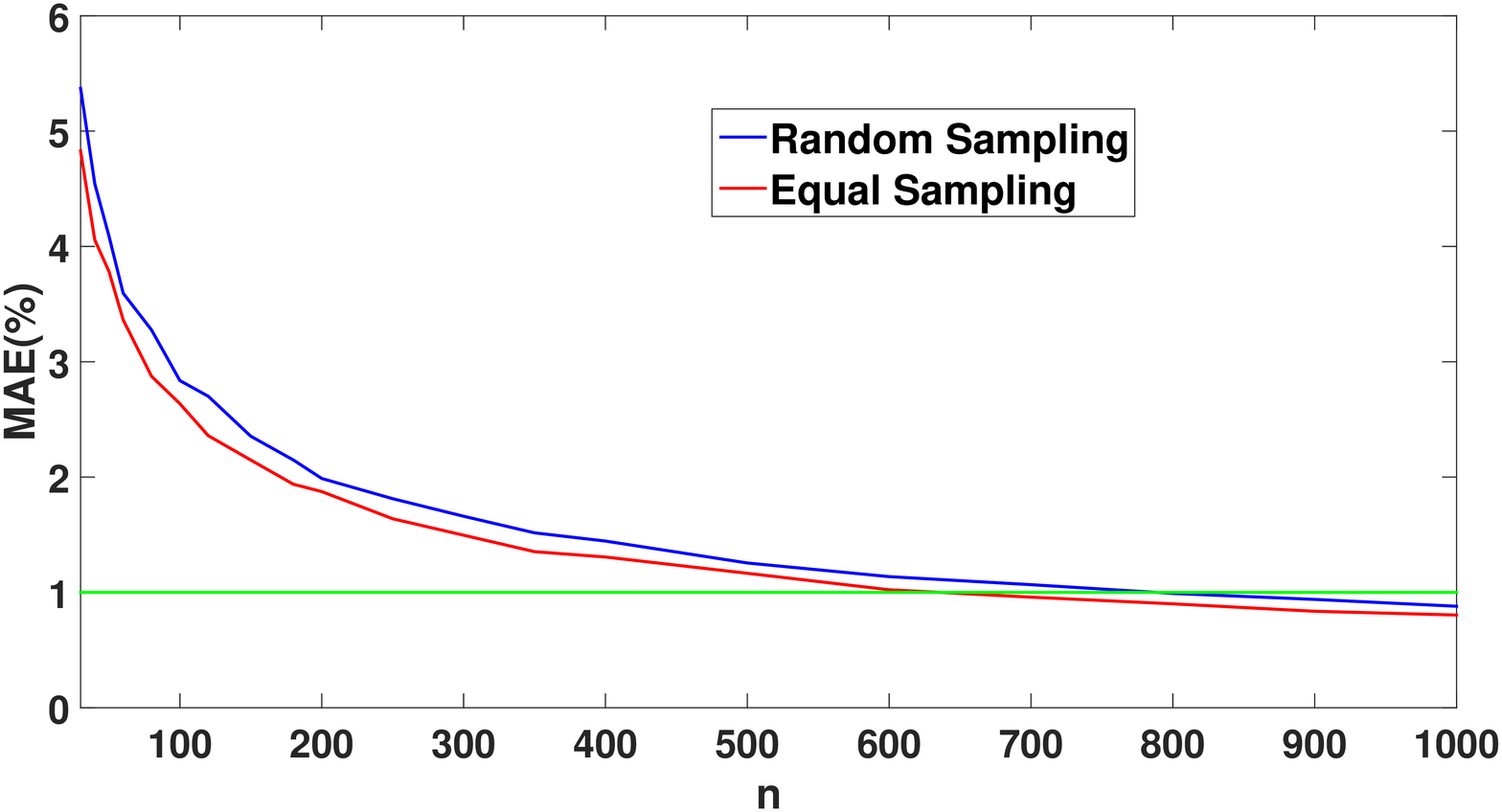}
    \caption{news20, $K=10$, SQRT}
    \label{fig:e3a}
  \end{subfigure}
  \begin{subfigure}[b]{0.49\columnwidth}
   \includegraphics[trim=0.8in 0.50in 1.0in 0.25in,width=1.0\columnwidth,height=1.0in]{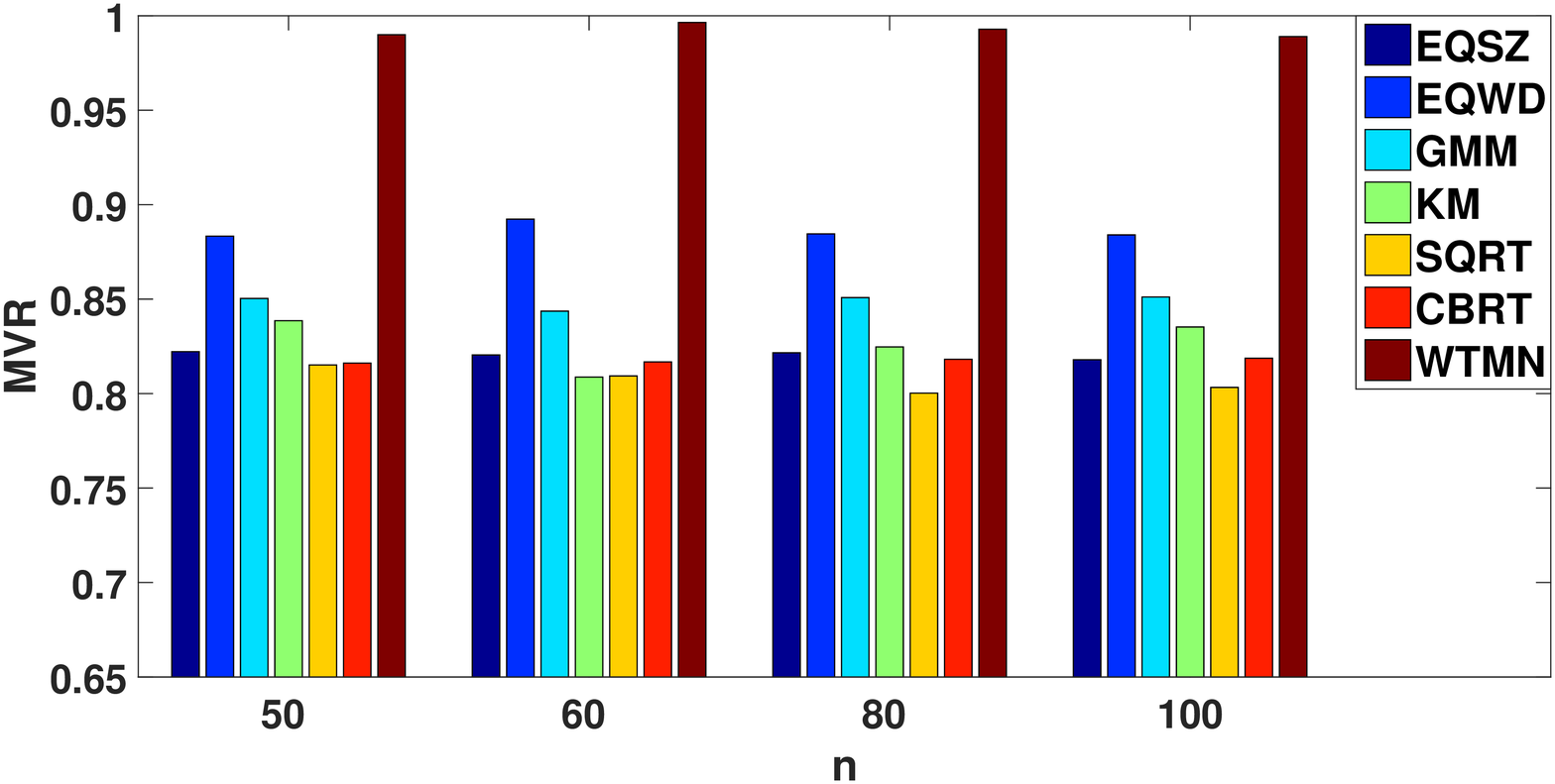}
    \caption{news20, $K=10$}
    \label{fig:e3b}
  \end{subfigure}
  \caption{Equal Stratified Sampling}
\end{figure}

The results on \emph{rcv1} dataset for Equal Allocation are shown is Figures \ref{fig:e1a} and \ref{fig:e1b}. Figure \ref{fig:e1a} uses KM based stratification. In this case $n$ required for $1\%$ error margin is reduced by a substantial amount which is close to $\mathbf{58.5\%}$ (from $284$ to $118$ ). Fig \ref{fig:e1b} shows that all stratification barring EQSZ and WTMN gives similar reduction in variance which is in the range of $\mathbf{55-60\%}$. Thus for \emph{rcv1} significant improvement in precision of accuracy estimates can be obtained using the Equal allocation policy. 

\begin{figure}[t]
  \begin{subfigure}[b]{0.49\columnwidth}
    \includegraphics[trim=1.0in 0.30in 0.7in 1.0in,width=1.0\columnwidth,height=1.0in]{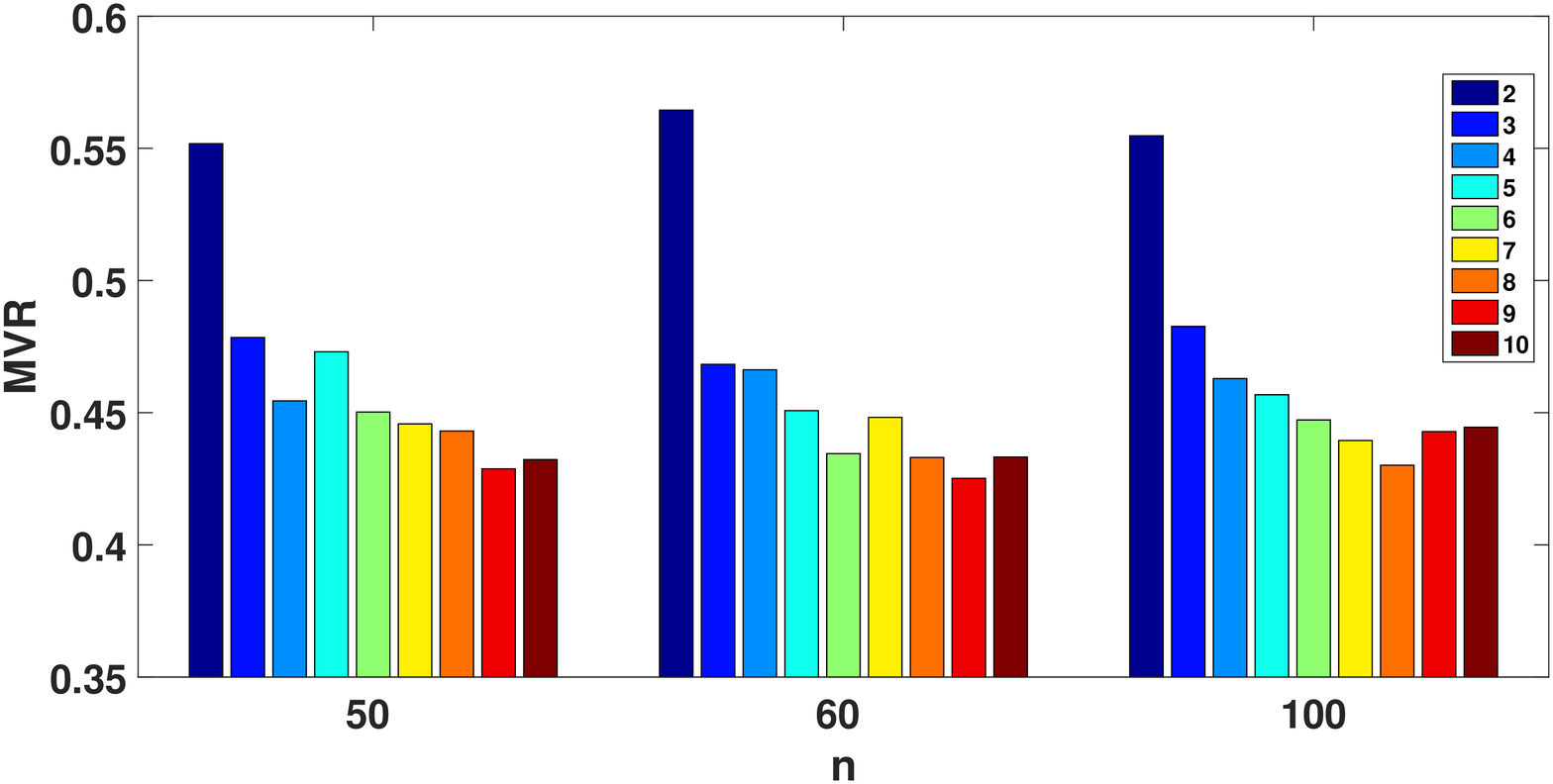}
    \caption{rcv1,Equal}
    \label{fig:ek1a}
  \end{subfigure}
  \begin{subfigure}[b]{0.49\columnwidth}
   \includegraphics[trim=1.0in 0.30in 1.0in 1.0in,width=1.0\columnwidth,height=1.0in]{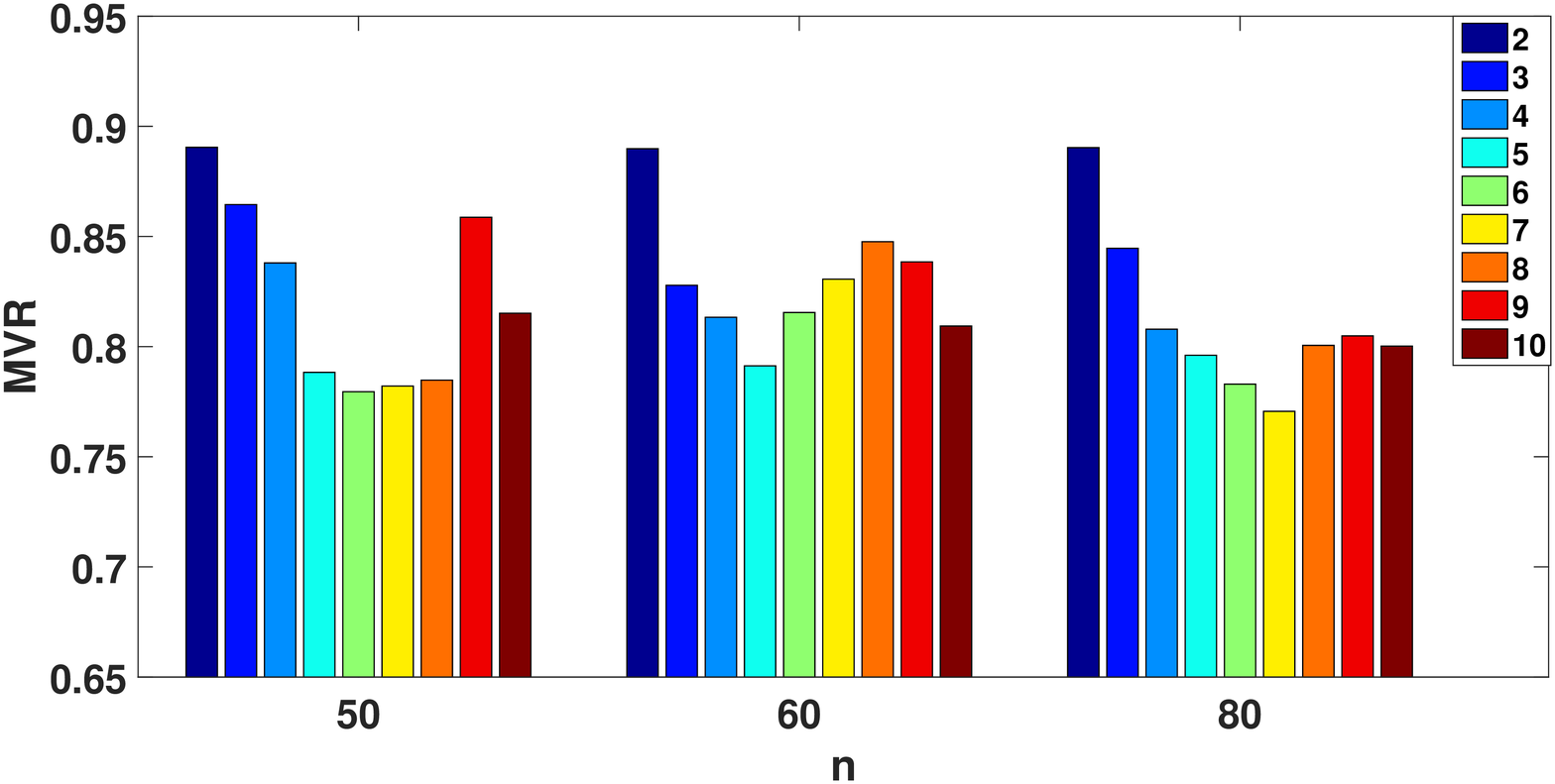}
    \caption{news20,Equal}
    \label{fig:ek1b}
  \end{subfigure}
  \caption{MVR Variation With K for Equal}
	\vspace{-0.05in}  
\end{figure}

Results on the \emph{epsilon} dataset are shown in Figures \ref{fig:e2a} and \ref{fig:e2b}. Fig \ref{fig:e2a} used EQSZ for stratification, resulting in about $16\%$ reduction in labeling resource for $1\%$ error. However, the more important point to be noted is that barring EQSZ all other stratification methods leads to an increase in variance of accuracy estimates compared to random sampling. This illustrates that equal allocation based stratified sampling does not come with the assurance that it will lead to reduction in an estimator's variance. The results for \emph{news20} dataset are shown in Fig \ref{fig:e3a} and \ref{fig:e3b}. In this case about $22\%$ reduction in labeling resource can be observed  and variation reduction lies in range of $18-23\%$ in most cases. The variation of MVR with $K$ for rcv1 and news20 is shown in Figure \ref{fig:ek1a} and \ref{fig:ek1b}. The trend is similar to what we observed as before.  

\subsection{Optimal Allocation}
In the previous section we observed that for \emph{rcv1} dataset Equal allocation resulted in a substantial reduction in the variance of the stratified accuracy estimator. The optimal allocation policy (OPT-A1 or OPT-A2) leads to further reduction in variance by only few percentage points ($4-6\%$ more) which does not translate into significant gain in terms of labeling resource reduction. It comes out to be slightly above $59\%$. This suggests that a data specific upper bound in variance reduction exists and equal allocation policy is not very far from this upper bound. This is clearly a data specific aspect as we observe that for \emph{epsilon} and \emph{news20} optimal allocation actually results in substantial reduction in variance. We use $n_{ini}=10$ for OPT-A1 algorithm. For OPT-A1 mid range $K$ such as $K=6\,\,or\,\,7$ are better in general, especially at lower $n$. $K$ affects the number of samples ($n_{ini}*K$) used up for initial estimation of $S_k$. Mid range $K$ are sufficient for obtaining good stratification and at the same time we are left with enough labeling resource which can be allocated optimally. 

\begin{figure}[t]
  \begin{subfigure}[b]{0.49\columnwidth}
    \includegraphics[trim=1.0in 0.0in 0.65in 0.0in,width=1.0\columnwidth,height=1.0in]{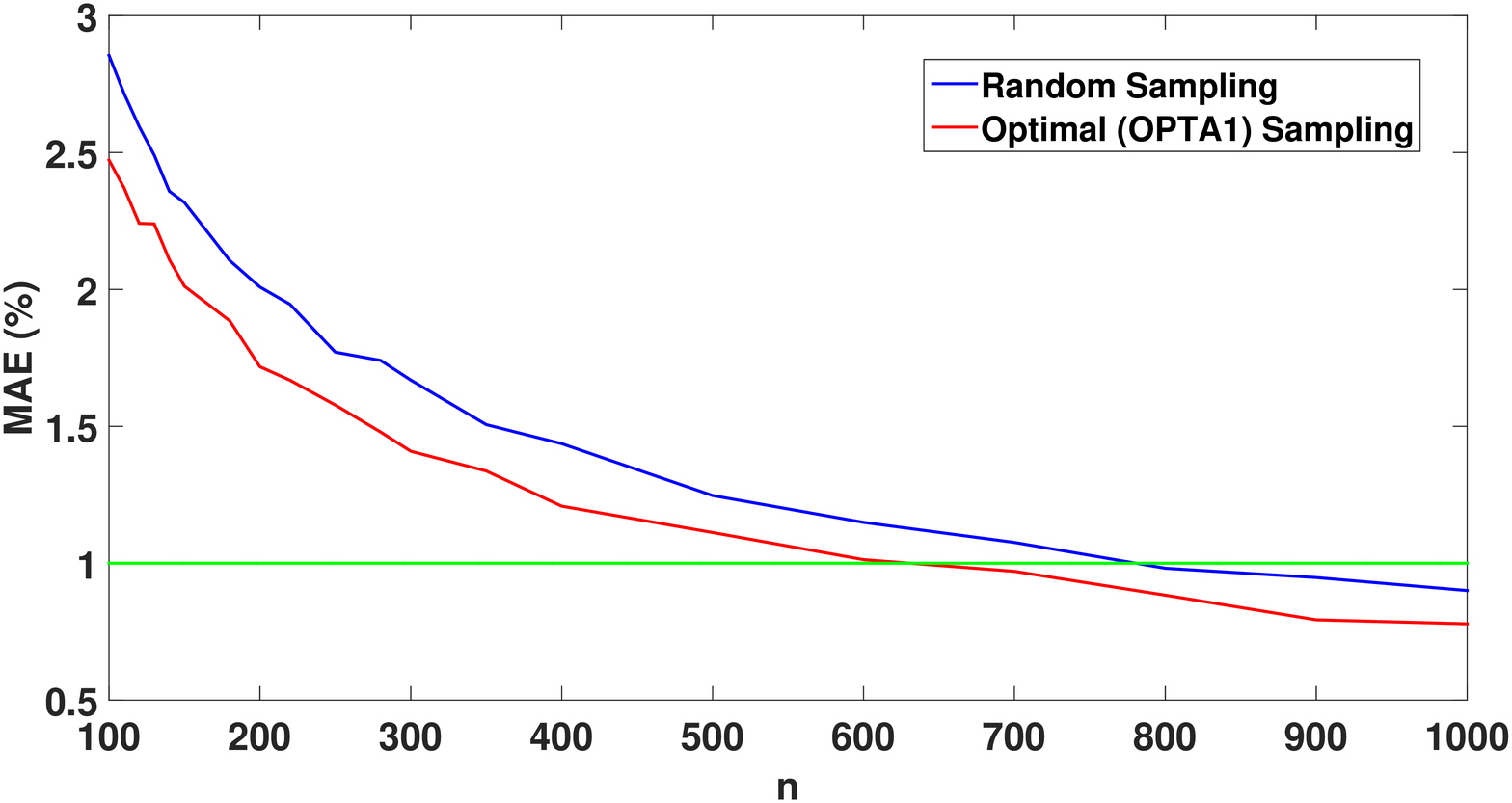}
    \caption{epsilon,EQSZ,$K=6$}
    \label{fig:oa12a}
  \end{subfigure}
  \begin{subfigure}[b]{0.49\columnwidth}
   \includegraphics[trim=1.0in 0.0in 1.0in 0.0in,width=1.0\columnwidth,height=1.0in]{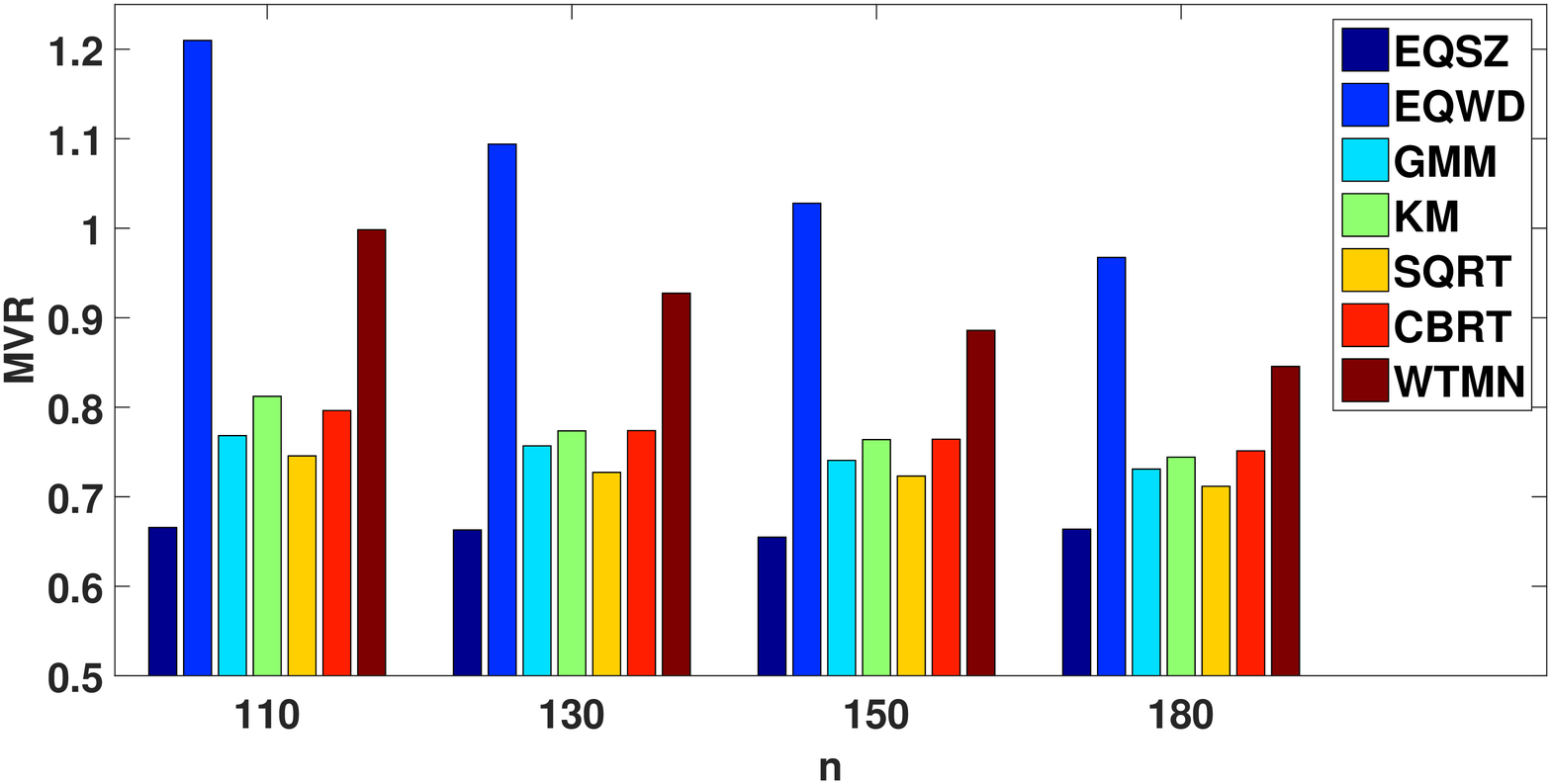}
    \caption{epsilon,$K=6$}
    \label{fig:oa12b}
  \end{subfigure}
    \begin{subfigure}[b]{0.49\columnwidth}
    \includegraphics[trim=1.0in 0.0in 0.65in 0.0in,width=1.0\columnwidth,height=1.0in]{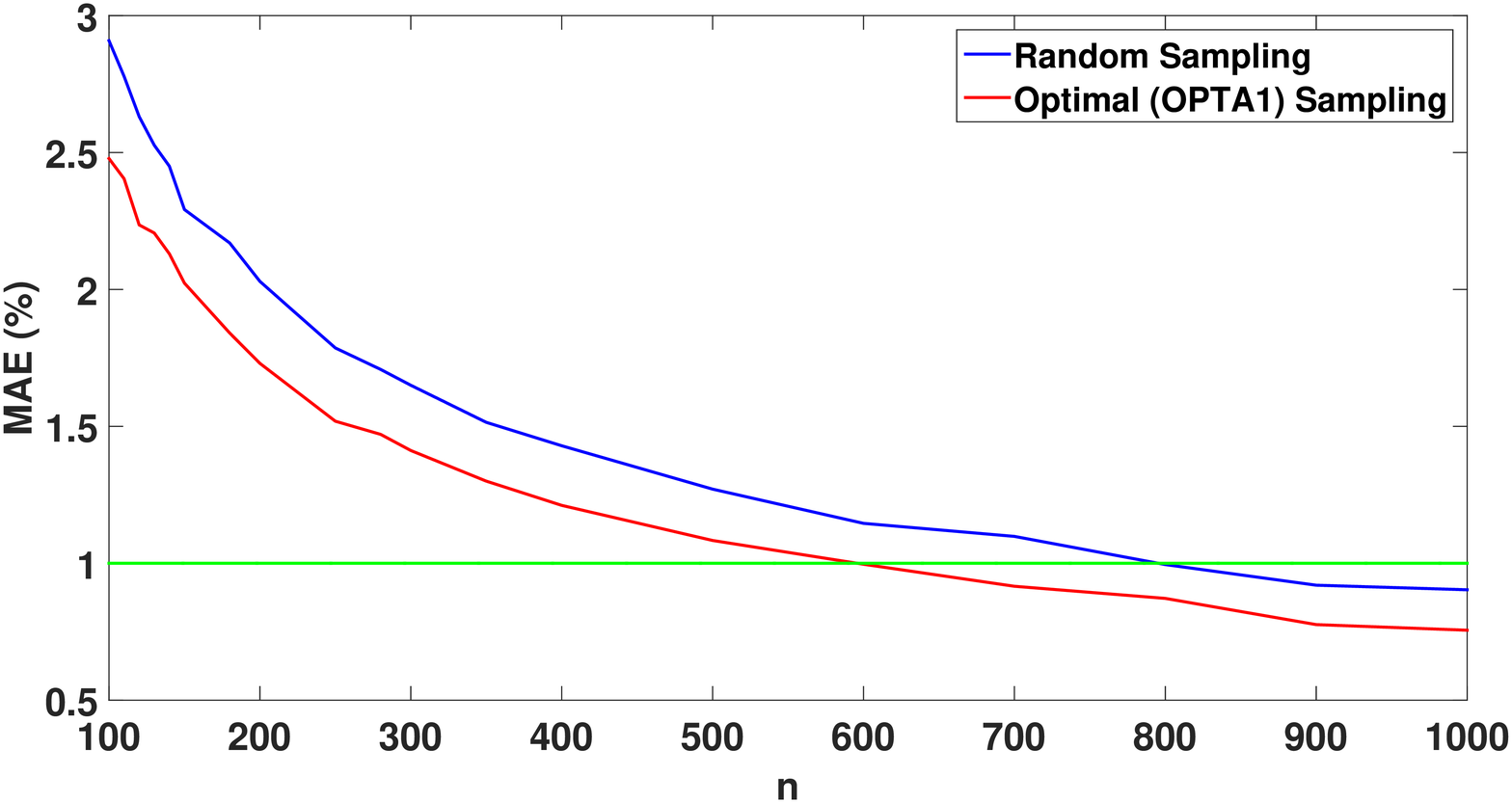}
    \caption{news20,$K=6$,SQRT}
    \label{fig:oa13a}
  \end{subfigure}
  \begin{subfigure}[b]{0.49\columnwidth}
   \includegraphics[trim=1.0in 0.0in 1.0in 0.0in,width=1.0\columnwidth,height=1.0in]{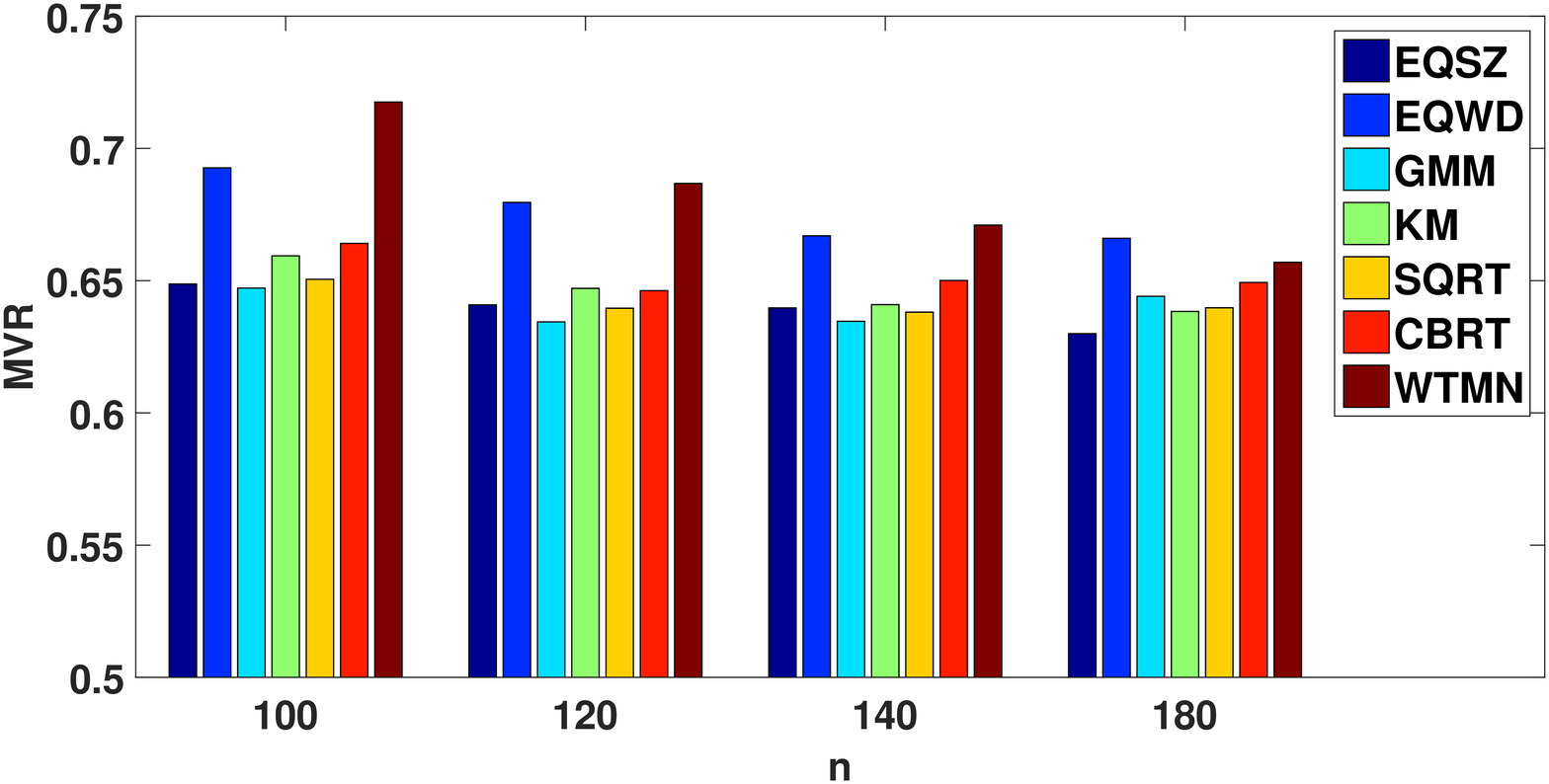}
    \caption{news20,$K=6$}
    \label{fig:oa13b}
  \end{subfigure}
  \caption{OPT-A1 Optimal Stratified Sampling}
  \vspace{-0.20in}
\end{figure}

Figure \ref{fig:oa12b} shows that EQSZ can results in over $\mathbf{30-35\%}$ reduction in variance compared to random sampling. In Figure \ref{fig:oa12a},  $n$ required for $1\%$ error is reduced by $23\%$ using OPT-A1 which is about $10\%$ and $7\%$ higher over proportional and equal allocation respectively. The worst stratification method is EQWD which corresponds to the practical problem we stated previously. Although, at higher $n$ it does lead to reduction in variance it is still not as good as other methods for stratification. For \emph{news20} OPT-A1 leads to reduction in $n$ by about $\mathbf{27\%}$ for $1\%$ error which is higher than that for proportional and equal by $11\%$ and $5\%$ respectively. The variance is reduced by more than $35\%$ for several cases which is substantially higher than other two allocation methods.

\subsubsection{OPT-A1 vs OPT-A2}
\begin{figure}[t]
  \begin{subfigure}[b]{0.49\columnwidth}
    \includegraphics[trim=1.0in 0in 0.7in 0in,width=1.0\columnwidth,height=1.2in]{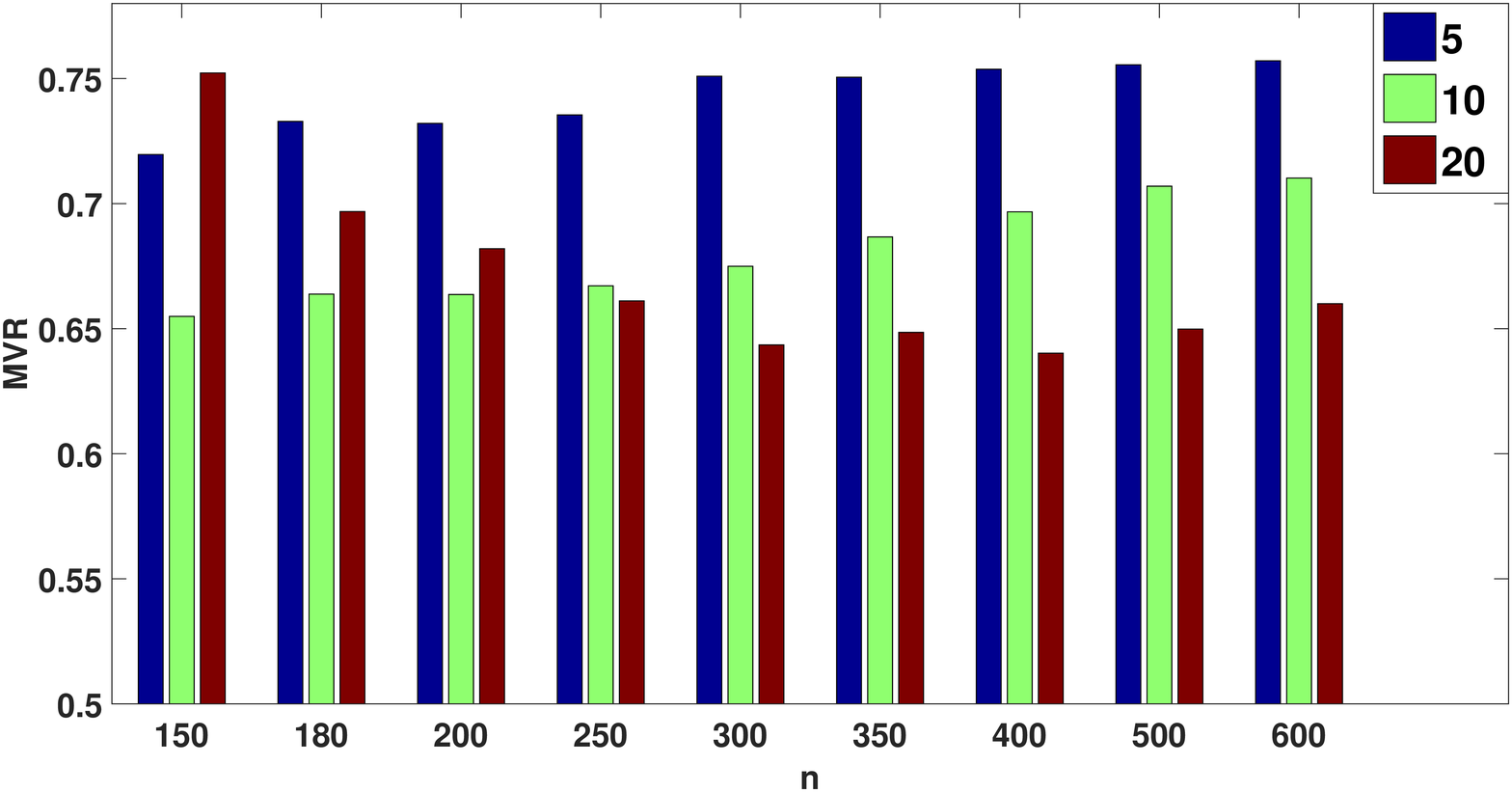}
    \caption{epsilon, EQSZ, $K=6$}
    \label{fig:init1}
  \end{subfigure}
  \begin{subfigure}[b]{0.49\columnwidth}
   \includegraphics[trim=1.0in 0in 1.0in 0in,width=1.0\columnwidth,height=1.2in]{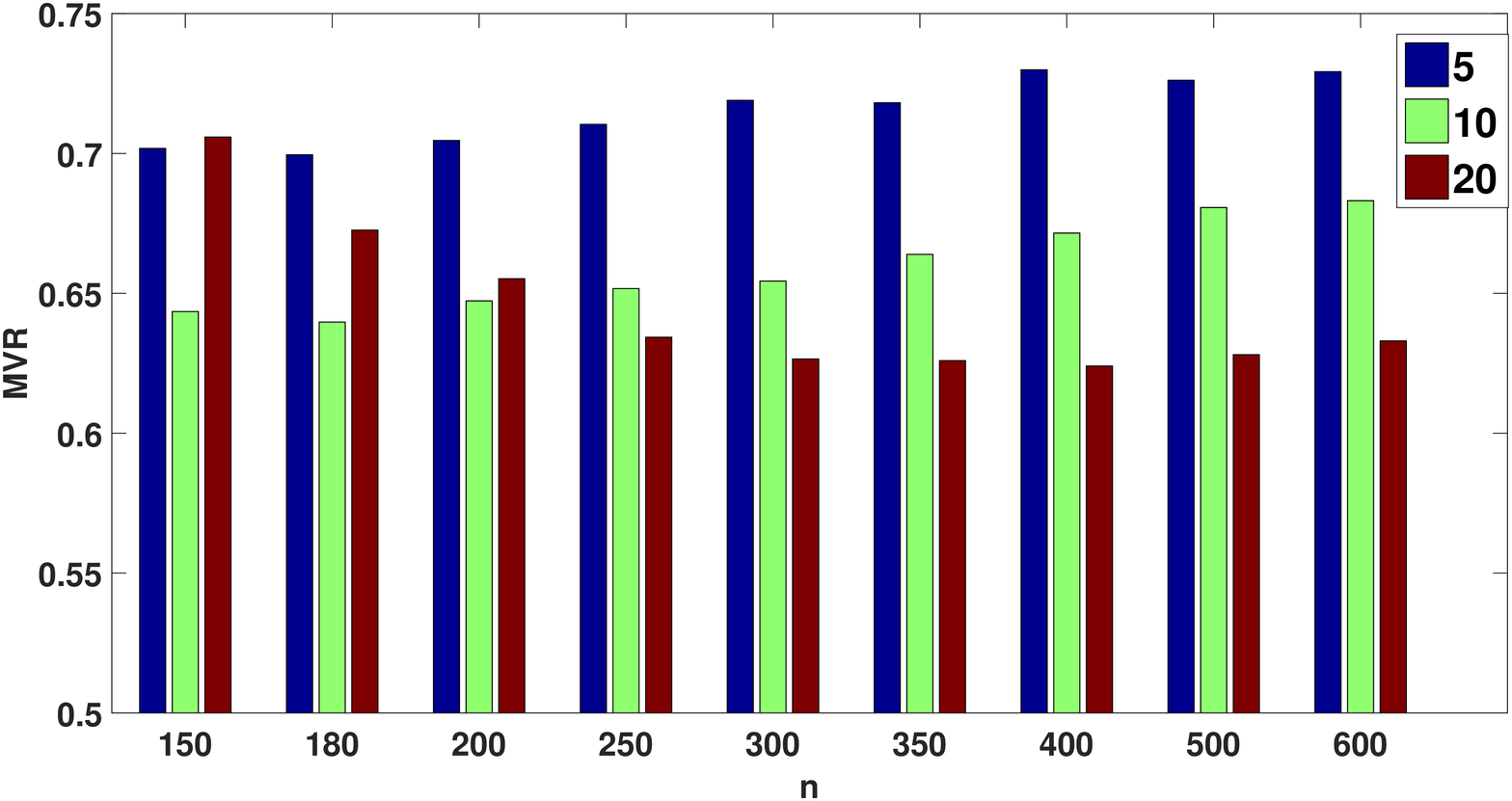}
    \caption{news20, SQRT, $K=6$}
    \label{fig:init2}
  \end{subfigure}
  \caption{OPT-A1 Dependence on $n_{ini}$}
  \vspace{-0.10in}
\end{figure}

We mentioned previously that setting the right $n_{ini}$ in OPT-A1 might present practical difficulties. This is illustrated in Figures \ref{fig:init1} and \ref{fig:init2} where we show MVR values for $n_{ini}$ equal to $5$,$10$ and $20$. We first observe that for sufficiently high $n$, higher $n_{ini}$ is better. This is expected as increasing $n_{ini}$ results in better estimation of $S_k$ and for large $n$ we are still left with enough labeling resource which can be allocated in an optimal sense to help achieve lower variance. However, the problem occurs for lower $n$ where we observe that MVR first reduces by increasing $n_{ini}$ from $5$ to $10$ but then increases substantially when we increase it further to $20$. Clearly, there is some optimal value between $5$ to $20$ which cannot be known a priori. 

\begin{figure}[t]
  \begin{subfigure}[b]{1.0\columnwidth}
    \includegraphics[trim=1.0in 0.0in 0.7in 0.0in,width=1.0\columnwidth,height=1.2in]{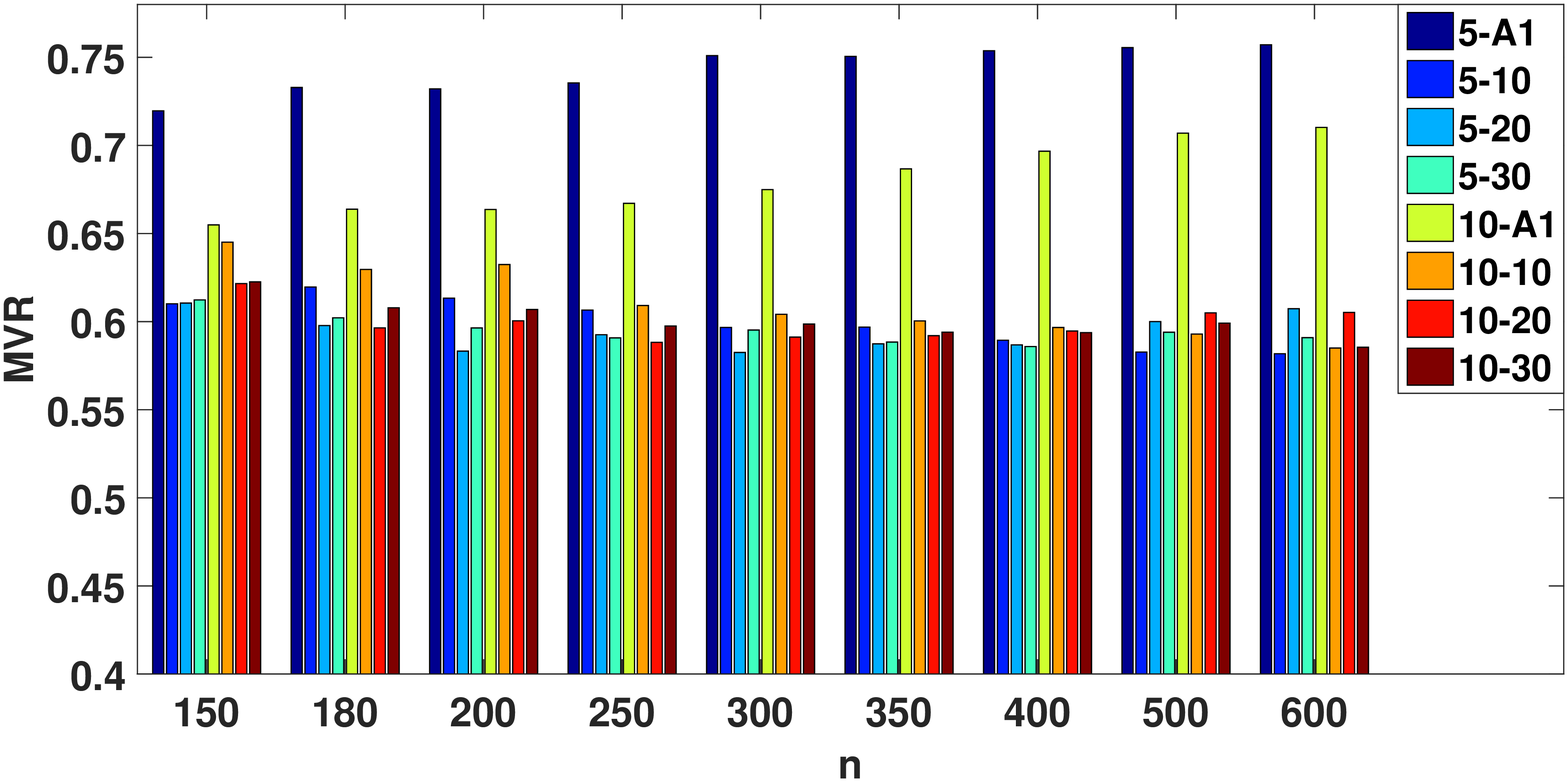}
    \caption{epsilon,EQSZ,$K=6$}
    \label{fig:oa21}
  \end{subfigure}
  \begin{subfigure}[b]{1.0\columnwidth}
   \includegraphics[trim=1.0in 0.0in 1.0in 0.0in,width=1.0\columnwidth,height=1.2in]{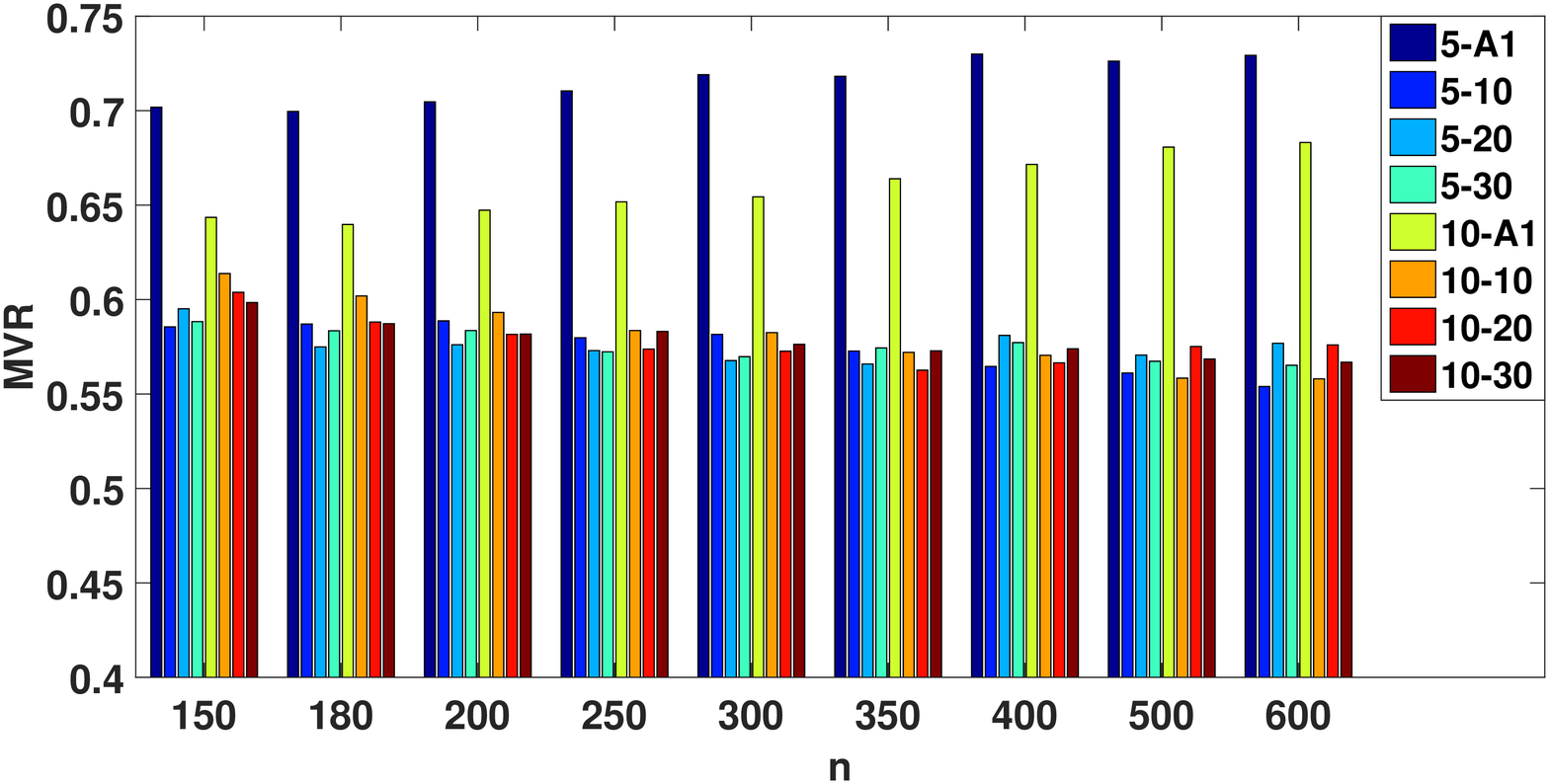}
    \caption{news20,SQRT,$K=6$}
    \label{fig:oa22}
  \end{subfigure}
  \caption{OPTA1 vs OPTA2, $n_{ini}$ and $n_{step}$}
  \vspace{-0.15in}
\end{figure}

To get around the problems of OPT-A1, we proposed OPT-A2. Figures \ref{fig:oa21} and \ref{fig:oa22} shows the efficiency and benefits of OPT-A2. For both figures legend are in form $n_{ini}-n_{step}$. $n_{ini}-A1$ legends represent the corresponding MVR using OPT-A1. First,  we observe that irrespective of the value of $n_{ini}$ OPT-A2 results in reduction of MVR. In comparison to OPT-A1, OPT-A2 leads to a further reduction in variance of estimated accuracy by upto $\mathbf{18\%}$ in certain cases. The range of reduction is $\mathbf{5-18\%}$.  This implies that for a given $n$, OPT-A2 will lead to a more precise estimation of true accuracy. Moreover, we observe that setting $n_{ini}$ is no more critical; $n_{ini}=5$ works as good as $n_{ini}=10$. Even more convenient is the fact that $n_{step}$ does not affect MVR in any major way which removes the role of any hyperparameter for OPT-A2. Hence, one can set $n_{ini}$ to any small value such as $5$ and any reasonable value of $n_{step}$  such as $10$ or $20$ works fine. As we mentioned before for \emph{rcv1} equal allocation OPT-A1 leads to only a small improvement in results over equal allocation. Using OPT-A2 on \emph{rcv1} dataset leads to a further small improvements in results over \emph{OPT-A1}, but not substantial. This again points toward existence of data specific bound. 
\begin{figure}[t]
    \includegraphics[trim=1.0in 0.0in 1.0in 0.0in,width=1.0\columnwidth,height=1.2in]{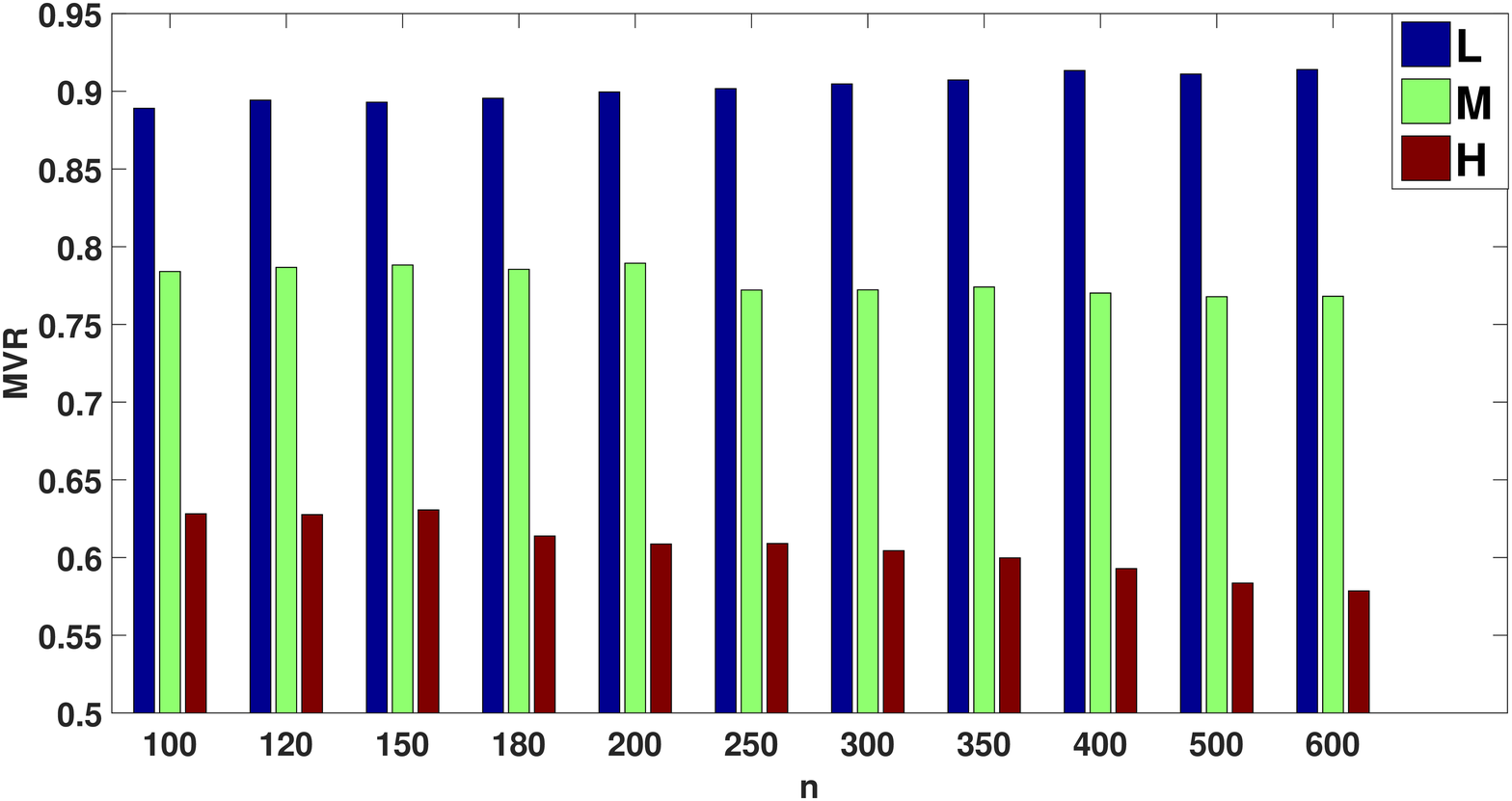}
    \caption{epsilon,EQSZ,$K=6$,OPTA2}
    \label{fig:accbsd}
    \vspace{-0.10in}
\end{figure}
\subsection{Dependence on True Accuracy}
It is expected that the value of true accuracy would have some effect on the MVR, which measures how well stratified sampling is doing compared to random sampling. Mainly, we would like to understand when can we expect MVR values to be low. In this section, we want to empirically study the effect of actual value of true accuracy on the proposed accuracy estimation process. For all three datasets, the accuracy of logistic regression and linear SVM are close and hence any reasonable analysis cannot be made by comparing performance for these two types of classifiers. 

We try to study this effect on the \emph{epsilon} dataset by training 3 different classifiers (SVMs) with varying accuracies. The true accuracies of the classifiers are $88\%$(H), $77\%$(M) and $67\%$(L). The classifier accuracy has been reduced by reducing the amount of training data used. Obviously, the test data $\mathcal{D}$ on which these accuracies have been computed is same for all 3 classifiers. Now we try to estimate these accuracies for the 3 classifiers by sampling from $\mathcal{D}$ and we observe the MVR values for different $n$. Figure \ref{fig:accbsd} show the results for three cases using OPT-A2 with $n_{ini}=5$ and $n_{step}=10$. We observe that MVR follows an inverse trend with classifier accuracy. Thus, the better the classifier the more effective stratified sampling is in reducing the variance of accuracy estimate. Similar trend for OPT-A1 also exist. 
\section{DISCUSSION AND CONCLUSIONS}
We presented a method for evaluating classifiers in a limited labeling budget scenario. We theoretically derived the variance of accuracy estimates for different cases and showed that \emph{stratified sampling} can be used for reducing the variance of accuracy estimates. We perform empirical study on both probabilistic (logistic regression) as well non-probabilistic (support vector machines) classifiers. We observed that in some cases this reduction can be as high as over $65\%$. This helps in obtaining a more precise estimate of accuracy for a given labeling budget or in other words reducing the labeling resource required to estimate accuracy with very low error. It is also worth noting that clustering methods in general perform as well as established stratification methods which are designed to approximate optimum point of stratification. One of the interesting outcomes is related to \emph{Equal} allocation. We noted that equal allocation which is much simpler to implement can work remarkably well in some cases. However, the downside is that it does not come with theoretical assurance that variance will always be reduced compared to random sampling and in an ill-structured stratification it might actually lead to increase in variance compared to simple random sampling. 

As far as optimal allocation is concerned we employed two methods for implementation. We showed that its implementation is best done through the proposed OPT-A2 method. Performance of OPT-A2 is not only better compared to OPT-A1 but is almost independent of the parameters ($n_{ini}$ and $n_{step}$) it takes as input. This is not the case for OPT-A1 where $n_{ini}$ plays a critical role. 

Stratified sampling seems to work well in reducing the variance of accuracy estimates when compared to random sampling, however, we found that its effectiveness decreases as the accuracy goes down. It remains to be seen whether stratification using the feature space of instances can address this problem or not. Other practical situations are cases when the distribution of labels in the test data is highly skewed. In this case simple random sampling might result in selection of only positive or negative data, especially if $n$ is very low. Evaluating classifiers based on only positive or negative data is not a desirable situation. Stratified sampling can be used to address this problem provided we factor in the skewness in the stratification step. We continue to investigate these problems of classifier evaluation.

\bibliographystyle{abbrv}
\bibliography{references}  
%
%

\end{document}